\documentclass[journal]{IEEEtran}
\usepackage{cite}
\usepackage{amsmath,amssymb,float, enumerate} 
\usepackage{algorithm}
\usepackage{times}
\usepackage{todonotes}

\newtheorem{theorem}{Theorem}
\newtheorem{corollary}{Corollary}

\newtheorem{lemma}{Lemma}
\newtheorem{example}{Example}

\def\hC{\widehat{C}}
\def\hc{\hat{c}}
\def\Y{{\cal Y}}

\def\ha{\hat{a}}

\def\hL{\widehat{L}}

\def\hp{\hat{p}}

\def\th{\hat{t}}
\def\hT{\widehat{T}}
\def\hx{\hat{x}}

\def\hw{\hat{w}}

\def\A{{\mathcal A}}

\def\D{{\mathcal D}}
\def\E{{\mathcal E}}

\def\L{{\mathcal L}}
\def\Q{{\mathcal Q}}
\def\T{{\mathcal T}}

\def\e{\epsilon}

\def\g{g}

\def\R{\mathbb{R}}

\def\I{\mathbb{I}}

\def\PROB{\mathbb P}
\def\P{{\PROB}}

\def\wh{\widehat}
\def\ol{\overline}
\def\at{@}

\newcommand{\bp}{\boldsymbol{p}}

\newlength{\minipagewidth}
\newcommand{\bookbox}[1]{
\par\bigskip\noindent
\begin{center}
\framebox[\minipagewidth]{
\begin{minipage}{0.9\minipagewidth}
{#1}
\end{minipage} }
\end{center}
\par\bigskip\noindent }

\allowdisplaybreaks

\begin{document}
\title{Efficient Tracking of Large Classes of Experts
\thanks{
This research was supported in part by the
National Development Agency of Hungary from the Research and Technological Innovation Fund
(KTIA-OTKA CNK 77782),
the Alberta Innovates Technology Futures,
the Natural Sciences and Engineering Research Council (NSERC) of
Canada, 
the Spanish Ministry of Science and Technology grant MTM2009-09063,
and the PASCAL2 Network of Excellence under EC grant no.\ 216886.}
\thanks{The material in this paper was presented in part at the 2012 IEEE International Symposium on Information Theory, Cambridge, MA, USA, July 2012.}
\thanks{
A. Gy\"orgy is with the Department of Computing Science, University of
Alberta, Edmonton, Alberta, Canada T6G 2E8; during part of this work he was
with the Machine Learning Research Group,
Computer and Automation Research Institute of the Hungarian
Academy of Sciences, Budapest, Hungary,
(email: {\tt gya{\at}szit.bme.hu}). 
T. Linder is with the Department of 
Mathematics and Statistics, Queen's University,
Kingston, Ontario, Canada K7L 3N6 (email: {\tt
linder{\at}mast.queensu.ca}). 
G. Lugosi is with ICREA and the Department of Economics, 
Pompeu Fabra University, Ramon Trias Fargas 25-27, 
08005 Barcelona, Spain (email: {\tt   gabor.lugosi@gmail.com}).
}
}

\author{Andr\'as Gy\"orgy,~\IEEEmembership{Member,~IEEE,} 
Tam\'as Linder,~\IEEEmembership{Senior Member,~IEEE,} and G\'abor Lugosi\\ \medskip \mbox{} \\
\today}

\maketitle

\begin{abstract}
In the framework of prediction of individual sequences, sequential
prediction methods are to be constructed that perform nearly as well
as the best expert from a given class. We consider prediction
strategies that compete with the class of switching strategies that
can segment a given sequence into several blocks, and follow the
advice of a different ``base'' expert in each block.  As usual, the
performance of the algorithm is measured by the regret defined as the
excess loss relative to the best switching strategy selected in
hindsight for the particular sequence to be predicted.  In this paper
we construct prediction strategies of low computational cost for the
case where the set of base experts is large. In particular we provide
a method that can transform any prediction algorithm $\A$ that is
designed for the base class into a tracking algorithm. The resulting
tracking algorithm can take advantage of the prediction performance and
potential computational efficiency of $\A$ in the sense that it can be
implemented with time and space complexity only $O(n^{\gamma} \ln n)$
times larger than that of $\A$, where $n$ is the time horizon and
$\gamma \ge 0$ is a parameter of the algorithm. With $\A$ properly
chosen, our algorithm achieves a regret bound of optimal order for
$\gamma>0$, and only $O(\ln n)$ times larger than the optimal order
for $\gamma=0$ for all typical regret bound types we examined.  For
example, for predicting binary sequences with switching parameters
under the logarithmic loss,
our method achieves the optimal $O(\ln n)$ regret rate with time
complexity $O(n^{1+\gamma}\ln n)$ for any $\gamma\in (0,1)$.
\end{abstract}

\section{Introduction}
\label{sec:intro}

In the on-line (sequential) decision problems considered in this
paper, a decision maker (or forecaster) chooses, at each time instant
$t=1,2,\ldots$, 
an action from a set. After each action taken,
the decision maker suffers some loss based on the state of the
environment and the chosen decision.  The general goal of the forecaster
is to minimize its cumulative loss. Specifically, the forecaster's aim
is to achieve a cumulative loss that is not much
larger than that of the 
best expert (forecaster) in a reference class $\E$, from which the
best expert is 
chosen in hindsight. This problem is known as ``prediction with expert
advice.'' The maximum excess loss $R_n$ of the forecaster relative to the best expert is called the (worst-case) cumulative regret, where the maximum is taken over all possible behaviors of the environment and $n$ denotes the time horizon of the problem.
Several methods are known that can compete successfully with different
expert classes in the sense that the regret only grows sub-linearly, that is, $\lim_{n\to\infty}R_n/n=0$.
We refer to \cite{CeLu06} for a
survey. 

While the goal in the standard online prediction problem is to perform
nearly as well as the best expert in the class $\E$, a more ambitious
goal is to compete with the best \emph{sequence} of expert predictions
that may switch its experts a certain, limited, number of
times. This, seemingly more complex, problem may be regarded as a
special case of the standard setup by introducing the so-called
\emph{meta experts}.  A meta expert is described by a sequence of base
experts $(i_1,\ldots,i_n)\in \E^n$, such that at time instants
$t=1,\ldots,n$ the meta expert follows the predictions of the ``base''
expert $i_t \in \E$ by predicting $f_{i_t,t}$.  The complexity of such
a meta expert may be measured by $C=|\{t\in\{1,2,\ldots,n-1\}: i_t\neq
i_{t+1}\}|$, the number of times it changes the base predictor (each
such change is called a switch).  Note that $C$ switches partition
$\{1,\ldots,n\}$ into $C+1$ contiguous segments, on each of which the
meta expert follows the predictions of the same base expert.  If a
maximum of $m$ changes are allowed and the set of base experts has $N$
elements, then the class of meta experts is of size $\sum_{j=0}^m
\binom{n-1}{j}N(N-1)^j$. Since the computational complexity of basic
prediction algorithms, such as the exponentially weighted average
forecaster, scales with the number of experts, a naive implementation
of these algorithms is not feasible in this case. However, several
more efficient algorithms have been proposed.

One approach, widely used in the information theory/source coding
literature, is based on transition diagrams \cite{Wil96,ShMe99}: A
transition diagram is used to define a prior distribution on the
switches of the experts, and the starting point of the current segment
is estimated using this prior. A transition diagram
defines a Markovian model on the switching times: a state of the model
describes the ``status'' of a switch process (corresponding to, e.g., the time when the last switch
occurred and the actual time), and the transition diagram defines the transition
probabilities among these states. In its straightforward version, at each
time instant $t$, the performance of an expert algorithm is emulated
for all possible segment starting points $1,\ldots,t$, and a weighted
average of the resulting estimates is used to form the next
prediction. In effect, this method converts an efficient algorithm to
compete with the best expert in a class $\E$ into one that competes
with the best sequence of experts with a limited number of
changes. The time complexity of the method depends on how complex the
prior distribution is, which determines the amount of computation
necessary to update the weights in the estimate. 
Note that a general prior distribution would require exponential
computational complexity in the sequence length, while at each time
instant the transition diagram model requires computations
proportional to the number of achievable states at that time instant. Using
a state space that describes the actual time, the time of the last
switch, and the number of switches so far, \cite{Wil96} provided a
prediction scheme achieving the optimal regret up to an additive constant (for the
logarithmic loss), and, omitting the number of switches from the
states, a prediction algorithm with optimal regret rate was provided.
\cite{ShMe99} showed (also for the logarithmic loss) that the transition probabilities in the latter
model can be selected so that the resulting prediction scheme achieves
the optimal regret rate with the best possible leading constant, and
the distributions they use allow
computing the weights at time instant $t$ with $O(t)$ complexity. As a
result, in $n$ time steps, the time complexity of the best
transition-diagram based algorithm is a factor $O(n)$ times larger
than that of the original algorithm that competes with $\E$, yielding
a total complexity that is quadratic in $n$.

For the same problem, a method of linear complexity was developed in
\cite{HeWa98}. It was shown in \cite{Vov99} that this method is equivalent to an easy-to-implement
weighting of the paths in the full transition diagram. Although, unlike transition diagram based methods, the original version of the algorithm of \cite{HeWa98} requires an a priori known upper bound on the
number of switches, the algorithm
can be modified to compete with meta experts with an arbitrary number
of switches: a linear complexity variant achieves this goal (by
letting its switching parameter $\alpha$ decrease to zero) at the
price of somewhat increasing the regret \cite{KoRo08}. A slightly
better regret bound can be achieved for the case when switching occurs
more often at the price of increasing the computational complexity
from linear to $O(n^{3/2})$ \cite{MoJa03,RoEr09} (by discretizing its
switching parameter $\alpha$ to $\sqrt{n}$ levels). 

In another approach, reduced transition diagrams have been used for the
logarithmic loss (i.e., lossless data compression) by \cite{WiKr97} and by
\cite{ShMe99} (the latter work considers a probabilistic setup as
opposed to the individual sequence setting). Reduced transition
diagrams are obtained by restricting some transitions, and
consequently, excluding some states from the original transition
diagram, resulting in (computationally) simpler models that, however,
have less descriptive power to represent switches.  An efficient
algorithm based on a reduced transition diagram for the general
tracking problem was given in \cite{GyLiLu08b}, while \cite{HaSe09}
developed independently a similar algorithm to minimize adaptive
regret, which is the maximal worst-case cumulative excess loss over
any contiguous time segment relative to a constant expert. It is easy
to see that algorithms with good adaptive regret also yield good
tracking regret.

An important  question is how one can compete with meta experts when
the base expert class $\E$ is very large.  In such cases special
algorithms are needed to compete with experts from the base class even
without switching. Such large base classes arise in on-line linear
optimization \cite{HaAgKa07}, lossless data compression
\cite{KrTr81,Sht87,WiShTj95}, the shortest path problem \cite{KaVe03,
TaWa03}, or limited-delay lossy data compression \cite{LiLu01, WeMe02,
GyLiLu04}.  Such special algorithms can easily be incorporated
in transition-diagram-based tracking methods, but the resulting
complexity is quadratic in $n$  (see, e.g.,  
\cite{ShMe99} for such an application to lossless data compression or
\cite{KoSi08,KoSi09,KoSi10} for applications to signal processing and universal
portfolio selection). If the special algorithms for large base
expert classes are combined with   the algorithm of \cite{HeWa98} to
compete with meta experts, the resulting algorithms  again have
quadratic complexity in $n$; see, e.g., \cite{Vov99,GyLiLu08} (the
main reason for this is that the special implementation tricks used for
the large base expert classes, such as dynamic programming, are incompatible
with the efficient implementation of the algorithm of \cite{HeWa98} for switching experts).
The only example we are aware of where efficient tracking algorithms
with linear time complexity are available for a meaningful, large class of base experts is
the case of online convex programming, where the set of base experts
is a finite dimensional convex set and the (time-varying) loss
functions are convex \cite{Zin03} (see also the related problem of tracking linear
predictors \cite{HeWa01}). In this case projected
gradient methods (including exponentially weighted average prediction)
lead to tracking regret bounds of optimal order.
Note that instead of the number of switches, these bounds measure the
complexity of the meta experts with the more refined notion of $L_p$ norms.

In this paper we tackle the complexity issue in competing with
meta-experts for large base expert classes by presenting a general
method for designing reduced transition diagrams. The resulting
algorithm converts any (black-box) prediction algorithm $\A$ achieving
good regret against the base-expert class into one that achieves good
tracking and adaptive regret. The advantage of this transition-diagram
based approach is that the conversion is independent of the base prediction algorithm $\A$,
and so some favorable properties of $\A$ are automatically transferred
to our algorithm. In particular, the complexity of our method depends on the
base-expert class only through the base prediction algorithm $\A$,
thus exploiting its potential computational
efficiency.\footnote{Other black-box reductions of forecasters
 for different notions of regret are available in the literature; for example, the conversion of
 forecasters achieving good external regret to ones achieving good
 internal regret \cite{StLu05,BlMa07}.}
Our algorithm
unifies and generalizes the algorithms of \cite{WiKr97, HaSe09} and
our earlier work \cite{GyLiLu08b}. This algorithm has an explicit
complexity-regret trade-off, covering essentially all such results in
the literature.  In addition to the (almost) linear complexity
algorithms in the aforementioned papers, the parameters of our
algorithm can be set to reproduce the methods based on the full
transition diagram \cite{Wil96, ShMe99, KoSi08}, or the
complexity-regret behavior of \cite{MoJa03,RoEr09}. Also, our
algorithm has regret of optimal order with complexity
$O(n^{1+\gamma}\ln n)$ for any $\gamma\in (0,1)$, while setting
$\gamma=0$ results in complexity $O(n \ln n)$ and a regret rate that is
only a factor of $\ln n$ larger than the optimal one (similarly to
\cite{WiKr97, GyLiLu08b, HaSe09}).

The rest of the paper is organized as follows. First the online
prediction and the tracking problems are introduced in
Section~\ref{sec_prelim}. In Section~\ref{sec:general} we describe our
general algorithm. Sections \ref{subsection:weight} and
\ref{sec_lowcomplex} present a unified method for the low-complexity
implementation of the general algorithm via reduced transition
diagrams.  Bounds for the performance the algorithm are developed in
Section~\ref{subsection:regret}. More explicit bounds are presented
for some important special cases in Sections \ref{subsection:exp} and
\ref{subsection:KT}. The results are extended to the related framework
of randomized prediction in Section~\ref{sec:random}. Some
applications to specific examples are given in
Section~\ref{sec:examples}.

\section{Preliminaries}
\label{sec_prelim}
In this section we review some basic facts about prediction with expert advice, and introduce the tracking problem.

\subsection{Prediction with expert advice}
\label{sec_pred}
Let the decision space
$\D$ be a convex subset of a vector space and let
$\Y$ be a set representing the outcome space.
Let $\ell:\D\times\Y \to \R$ be a loss function, assumed to be convex
in its first argument.
At each time instant $t=1,\ldots,n$, the environment chooses
an action $y_t\in\Y$ and each ``expert'' $i$ from a reference class $\E$ forms its prediction 
$f_{i,t} \in \D$.  
Then the forecaster chooses an action $\wh{p}_t \in\D$ (without knowing $y_t$),
suffers loss $\ell(\wh{p}_t,y_t)$, and the losses
$\ell(f_{i,t},y_t), i \in \E$ are revealed to the
forecaster.
(This is known as the full information case and in this paper
we only consider this model. In other, well-studied, variants
of the problem, the forecaster only receives limited information
about the losses.)

\begin{figure}
\bookbox{
\begin{center}
PREDICTION WITH EXPERT ADVICE
\end{center}
\noindent
For each round $t=1,2,\ldots$

\smallskip\noindent
\begin{itemize}
\item[(1)] 
the environment chooses the next outcome $y_t$ and the expert advice $\{f_{i,t}\in\D:i\in\E\}$;
the expert advice is revealed to the forecaster;
\item[(2)] 
the forecaster chooses the prediction $\wh{p}_t\in\D$;
\item[(3)]
the environment reveals the next outcome $y_t\in\Y$;
\item[(4)]
the forecaster incurs loss $\ell(\wh{p}_t,y_t)$ and each expert $i$ incurs loss $\ell(f_{i,t},y_t)$.
\end{itemize}
}
\caption{The repeated game of prediction with expert advice.}
\label{fig1}
\end{figure}

The goal of the forecaster is to minimize its cumulative loss
$\hL_n=\sum_{t=1}^n\ell(\wh{p}_t,y_t)$, which is equivalent to
minimizing its excess loss $\hL_n - \min_{i\in\E} L_{i,n}$ relative
to the the set of experts $\E$, where $L_{i,n}=\sum_{t=1}^n
\ell(f_{i,t},y_t)$ for all $i\in\E$.

Several methods are known that can compete successfully with different
expert classes $\E$ in the sense that the (worst-case) cumulative regret, defined as 
\begin{eqnarray*}
R_n&=& \max_{(y_1,\ldots,y_n) \in \Y^n} \left( \hL_n - \min_{i\in\E}  L_{i,n} \right) \\
&=& \max_{(y_1,\ldots,y_n) \in \Y^n} \left(\sum_{t=1}^n\ell(\wh{p}_t,y_t) - \min_{i\in\E} \sum_{t=1}^n \ell(f_{i,t},y_t) \right)
\end{eqnarray*}
only grows sub-linearly, that is, $\lim_{n\to\infty}R_n/n=0$.  
One of the most popular among these
is  \emph{exponential weighting}. 
When the expert class $\E$ is finite or countably infinite,
this method assigns, at each time instant
$t$,  the nonnegative weight 
\[
\pi_{i,t} =\frac{w_i e^{-\eta_t L_{i,t-1}}}{\sum_{j\in \E} w_j
  e^{-\eta_t L_{i,t-1}}}
\]
to each expert $i \in \E$. Here 
$L_{i,t-1}=\sum_{s=1}^{t-1} \ell(f_{i,s}, y_s)$ is the
cumulative loss of expert $i$ up to time $t-1$, $\eta_t>0$ is called the
learning parameter, and the  $w_i >0$ are  nonnegative initial weights with
$\sum_{i\in \E} w_i=1$, so that $\sum_{i \in \E} \pi_{i,t}=1$  (we define $L_{i,0}=0$ for all $i\in\E$, as well
as $\hL_0=0$). The
decision chosen by this algorithm is
\begin{equation}
\label{expweight}
\wh{p}_t=\sum_{i \in \E} \pi_{i,t} f_{i,t}
\end{equation}
which is well defined since $\mathcal{D}$ is convex.

In this paper we concentrate on two special types of loss functions:
bounded convex and exp-concave. For such loss functions the regret of
the exponentially weighted average forecaster is well understood. For
example, assume $\ell$ is convex in its first argument and takes its
values in $[0,1]$, and the set of experts is finite with $|\E|=N$. If
$\eta_t$ is nonincreasing in $t$, then for all $n$,
\begin{equation}
\label{convex2}
\hL_n \le \min_{i} \left\{ L_{i,n} + \frac{1}{\eta_n}\ln \frac{1}{w_i} \right\}+\sum_{t=1}^n \frac{\eta_t}{8}~,
\end{equation}
see \cite{ChZh10}.
By setting the initial weights to $w_i=1/N, i=1,\ldots,N$ and with  the choice
$\eta_t=2\sqrt{\ln N/t}$, one obtains for all $n \ge 1$,
\begin{equation}
\label{convex}
R_n \le \sqrt{n\ln N}~. 
\end{equation}
If, on the other hand, for some $\eta>0$ the function
$F(p)=e^{-\eta\ell(p,y)}$ is concave for any fixed $y\in \Y$ (such
loss functions are called \emph{exp-concave}) then, choosing $\eta_t
\equiv \eta$ and $w_i=1/N, i=1,\ldots,N$, one has   for all $n \ge 1$,
\begin{equation}
\label{expconcave}
R_n \le \frac{\ln N}{\eta}~.
\end{equation}
We note that the regret bounds in \eqref{convex2}--\eqref{expconcave}
do not require a fixed time horizon, that is, they hold simultaneously for all
$n\ge 1$. 

The family of exp-concave loss functions includes, for
example, for $p,y \in [0,1]$, the square loss $\ell(p,y)=(p-y)^2$ with
$\eta \le 1/2$, and the relative entropy loss
$\ell(p,y)=y\ln\frac{y}{p}+(1-y)\ln\frac{1-y}{1-p}$ with $\eta\le
1$. A special case of the latter is the logarithmic loss defined for
$y \in \{0,1\}$ and $p\in [0, 1]$ by $\ell(p,y)=-\I_{y=1}\ln p -
\I_{y=0} \ln (1-p)$, which plays a central role in data
compression. Here and throughout the paper $\I_B$ denotes the
indicator of event $B$.  We refer to \cite{CeLu06} for discussions of
these bounds.

\subsection{The tracking problem}

In the standard online prediction problem the goal is to perform as well as the best expert in a given reference class $\E$. In this paper we consider the more ambitious goal of competing with a sequence of expert predictions that are allowed to switch between experts. Formally, such a \emph{meta expert} is defined as follows.
Fix the time horizon $n\ge 1$. A meta expert that changes base experts
at most $C \ge 0$ times can be described by a vector of experts
$a=(i_0,\ldots,i_{C})\in \E^{C+1}$ and a ``transition path''
$T=(t_1,\ldots,t_C;n)$ such that $t_0:=1<t_1<\ldots<t_C<t_{C+1}:=n+1$.
For each $c=0,\ldots,C$, the meta expert follows the advice of expert
$i_c$ in the time interval $[t_c,t_{c+1})$. When the time horizon $n$
is clear from the context, we will omit it from the description of
$T$ and simply write $T=(t_1,\ldots,t_C)$. We note that this
representation is not unique as the definition does not require that
base experts $i_c$ and $i_{c+1}$ be different. Any meta expert that
can be defined using a given transition path $T$ is said to follow $T$.

The total loss of the meta expert   indexed by $(T,a)$, accumulated
during $n$ rounds, is 
\[
  L_n(T,a) = \sum_{c=0}^C L_{i_c}(t_c,t_{c+1})
\]
where $L_i(t_1,t_2)=\sum_{t=t_1}^{t_2-1}\ell(f_{i,t},y_t)$ denotes the loss of expert $i \in \E$ in the interval $[t_1,t_2), 1 \le t_1 \le t_2 \le n$.
For any $t \ge 1$, let $\T_t$ denote the set of all transition paths
up to time $t$ represented by vectors $(t_1,\ldots,t_C;t)$ with $1<
t_1 < t_2 < \ldots < t_C\le t$ and $0 \le C\le t$. For any
$T=(t_1,\ldots,t_C) \in \T_n$ and $t\le n$ define the truncation of
$T$ at time $t$ as $T_t=(t_1,\ldots,t_k; t)$, where $k$ is such that
$t_k \le t < t_{k+1}$ (note that $t \le n$ guarantees that $t_{C+1}=n+1>t$, and so $t_{k+1}$ is well-defined).  Furthermore, let $\tau_t(T)=\tau_t(T_t)=t_k$
denote the last change up to time $t$, and let $C_t(T)=C(T_t)=k$
denote the number of switches up to time $t$.  A transition path $T$
with $C$ switches splits the time interval $[1,n]$ into $C+1$
contiguous segments.

Our goal is to perform nearly as well as the meta-experts, that is, to
keep the regret $\hL_n - L_n(T,a)$ small relative to the meta-experts
$(T,a)$ for all outcome sequences $y_1,\ldots,y_n$. It is clear that
this cannot be done uniformly well for all meta experts; for example,
it is obvious that the performance of a meta expert that is allowed to
switch experts at each time instant cannot be achieved for all outcome
sequences. Indeed, it is known \cite{Mer93,HeWa98} that, for
exp-concave loss functions, the worst-case regret of any prediction
algorithm relative to the best meta-expert with at most $C$ switches,
selected in hindsight, is at least of the order of
$(C+1)\log n$,
where the worst-case tracking regret with respect to meta experts with
at most $C$ switches is defined as
\[
\max_{y_1,\ldots,y_n} \left( \hL_n - \min_{(T,a):C_n(T)=C} L(T,a)\right).
\]
Algorithms achieving optimal regret rates are known under general
conditions: for general convex loss functions and a finite number of
base experts, a tracking regret of order   
$(C(T)+1)\sqrt{n\ln n}$ (or $\sqrt{\smash[b]{(C+1) n \ln n}}$ if 
$C$ is known in advance) can be achieved \cite{HeWa98,Vov99,GyLiLu08}, 
while the $O((C+1)\ln n)$ lower bound is achievable in case of
exp-concave loss functions and a finite number of experts \cite{Wil96,
  HeWa98,ShMe99, KoSi08, KoRo08}, or when the base experts form a convex
subset of a finite dimensional linear space \cite{HaSe07}.

We will also consider the related notion of \emph{adaptive regret}
\[
R^a_n= \max_{t \le t'} \max_{y_t,y_{t+1},\ldots,y_{t'}} 
\left(\sum_{\tau=t}^{t'} \ell(\hp_\tau,y_\tau) - \min_{i \in
  \E}\sum_{\tau=t}^{t'}  \ell(f_{i,\tau},y_\tau) \right)
\]
introduced in  \cite{HaSe07} and \cite{HaSe09}, which is the maximal worst-case
cumulative excess loss over any contiguous time segment relative to a
constant expert. Minimizing the tracking and the adaptive regret are
similar problems.  In fact, one can show that the FLH1 algorithm
of \cite{HaSe07} developed to minimize the adaptive regret and a
dynamic version of the fixed-share algorithm of \cite{HeWa98}
introduced by \cite{KoRo08} to minimize the tracking regret are
identical. Furthermore, any algorithm with small adaptive regret also
enjoys small tracking regret, since the regret, in $n$ time steps,
relative to a meta expert that can switch the base expert $C$ times
can be bounded by $(C+1)R_n^a$. Although tracking regret bounds do not
immediately yield bounds on the adaptive regret (since the regret on a
time segment may be negative), it is usually straightforward to
modify the proofs for tracking regret to obtain bounds on the adaptive
regret; see, e.g., the proof of Theorem~\ref{thm:main}.

\section{A reduced complexity tracking algorithm}
\label{sec:track}

\subsection{A general tracking algorithm}
\label{sec:general}

Here we introduce a general tracking method which forms the basis of
our reduced complexity tracking algorithm.  Consider an on-line
forecasting algorithm $\A$ that chooses an element of the decision
space depending on the past outcomes and the expert advices according
to the protocol described in Figure \ref{fig1}.  Suppose that for all
$n$ and possible  outcome sequences of length $n$, $\A$ satisfies a regret bound
\begin{equation}
\label{eq:univbound}
R_n \le \rho_\E(n)
\end{equation}
with respect to the base expert class $\E$, where
$\rho_\E:[0,\infty)\to [0,\infty)$ is a nondecreasing
and concave function with $\rho_\E(0)=0$. These assumptions on $\rho_\E$ are
usually satisfied by the known regret bounds for different algorithms, such 
as the bounds \eqref{convex} and \eqref{expconcave} (with defining $\rho_\E(0)=0$ in the latter case). 
Suppose $1\le t_1 < t_2\le n$ and an instance of $\A$ is used for time instants 
$t \in [t_1,t_2):= \{t_1,\ldots,t_2-1\}$, that is, algorithm $\A$
is run  on data obtained in the segment $[t_1,t_2)$. The accumulated
loss of $\A$
during this period will be denoted by $L_\A(t_1,t_2)$. Then \eqref{eq:univbound} implies
\[
L_\A(t_1,t_2) -\min_{i\in\E} L_i(t_1,t_2) \le \rho_\E(t_2-t_1).
\]

Running algorithm $\A$ on a transition path $T=(t_1,\ldots,t_C;n)$ means that
at the beginning of each segment of $T$ (at time instants $t_c$) we restart
$\A$; this algorithm will be denoted in the sequel by $(\A,T)$. Denote
the output of this algorithm at time $t$ by
$f_{\A,t}(T_t)=f_{\A,t}(\tau_t(T))$. This notation emphasizes the fact
that, since $\A$ is restarted at the beginning of each segment of $T$,
the output of $(\A,T)$ at time $t$ is influenced by $T$ only through $\tau_t(T)$, the beginning of
the segment that includes $t$. The loss of  algorithm  $(\A,T)$ up to
time $n$ is
\[
L_n(\A,T)=\sum_{c=0}^C L_\A(t_c,t_{c+1})~.
\]

As most tracking algorithms, our algorithm will use weight functions
$w_t:\T_t\to[0,1]$ 
satisfying 
\begin{equation}
\label{eq:wcond}
\sum_{T_t \in\T_t} \!\!w_t(T_t)=1 \text{ and }
w_t(T_t)=\!\!\!\sum_{T'_{t+1} \in \T_{t+1}: T'_t=T_t} \!\!\!w_{t+1}(T'_{t+1})
\end{equation}
for all $t=1,2,\ldots$ and $T \in\T$.
Thus each $w_t$ is a probability distribution on $\T_t$ such that the
family $\{w_t; t=1,\ldots,n\}$ is consistent.  To simplify the notation,
we formally define $T_0$ as the ``empty transition path''
$\T_0:=\{T_0\}$, $L_0(\A,T_0):=0$, and $w_0(T_0):=1$.

We say that $\hT \in \T_n$ \emph{covers} $T\in \T_n$ if
the change points of $T$ are also change points of $\hT$.  Note that if
$\hT$ covers $T$, then any meta expert that follows transition path $T$
also follows transition path $\hT$.  We say that $w_n$ \emph{covers}
$\T_n$ if for any $T\in \T_n$ there exists a $\hT \in \T_n$ with
$w_n(\hT)>0$  which covers $T$.

Now we are ready to define our first master algorithm, given in
Algorithm~\ref{alg:master}. We note that the consistency of $\{w_t\}$
implies that, for any time horizon $n$, Algorithm~\ref{alg:master} is
equivalent to the exponentially weighted average forecaster
(\ref{expweight}) with the set of experts $\{(\A,T): T \in \T_n,
w_n(T_n)>0\}$ and initial weights $w_n(T)$ for $(\A,T)$. The
performance and the computational complexity of the algorithm heavily
depend on the properties of $w_t$; in this paper we will concentrate
on judicious  choices of $w_t$ that allow efficient computation of the
summations in Algorithm~\ref{alg:master} and have good prediction
performance.

\begin{algorithm}[!h]
\caption{General tracking algorithm.}
\label{alg:master}
{\bf Input:} prediction algorithm $\A$, weight functions $\{w_t; t=1,\ldots,n\}$, 
learning parameters $\eta_t>0, t=1,\ldots,n$. \\
For $t=1,\ldots,n$ predict 
\[
\wh{p}_t=\frac{\sum_{T \in \T_t} w_t(T) e^{-\eta_t L_{t-1}(\A,T_{t-1})} f_{\A,t}(\tau_t(T))}
              {\sum_{T \in \T_t} w_t(T) e^{-\eta_t L_{t-1}(\A,T_{t-1})}}~.
\] 
\end{algorithm}

The next lemma gives an upper bound on the performance of
Algorithm~\ref{alg:master}.  

\smallskip

\begin{lemma}
\label{lem:master}
Suppose $\eta_{t+1} \le \eta_t$ for all $t=1,\ldots,n-1$, the transition path $T_n$ is covered by $\hT_n=(\th_1,\ldots,\th_{C(\hT_n)})$ such that
$w_n(\hT_n)>0$, and $\A$ satisfies the regret bound \eqref{eq:univbound}.
Assume that the loss function $\ell$ is convex in its first argument and
takes values in the interval $[0,1]$.
Then for any meta expert $(T_n,a)$,
the regret of Algorithm~\ref{alg:master} is bounded as
\begin{eqnarray}
\lefteqn{\hL_n -  L_n(T_n,a)} \nonumber \\
 &\le & \sum_{c=0}^{C(\hT_n)} \rho_\E(\th_{c+1}-\th_c) +\sum_{t=1}^n
\frac{\eta_t}{8} + \frac{1}{\eta_n} \ln \frac{1}{w_n(\hT_n)} \nonumber \\ 
&\le &
(C(\hT_n)+1)\rho_\E\!\!\left(\frac{n}{C(\hT_n)+1}\right)\nonumber \\
& & \mbox{}  +\sum_{t=1}^n \frac{\eta_t}{8} + \frac{1}{\eta_n} \ln \frac{1}{w_n(\hT_n)}~.
\label{eq:lemconvex}
\end{eqnarray}
On the other hand, if $\ell$ is exp-concave for the value of $\eta$ and
Algorithm~\ref{alg:master} is used with $\eta_t \equiv \eta$, then
\begin{eqnarray}
\lefteqn{\hL_n -  L_n(T_n,a)} \quad  \nonumber \\
&\! \!\! \!\!\!\!\!\! \le  &\!\!  \!\!\!   \sum_{c=0}^{C(\hT_n)} \rho_\E(\th_{c+1}-\th_c) + \frac{1}{\eta}
\ln \frac{1}{w_n(\hT_n)} \nonumber \\
& \! \!\!\!\!\!\! \!\!\le & \! \! \!\!\!  (C(\hT_n)+1)\rho_\E\left(\frac{n}{C(\hT_n)+1}\right) +
\frac{1}{\eta} \ln \frac{1}{w_n(\hT_n)}~. \label{eq:lemexpconcave}
\end{eqnarray}
\end{lemma}
\begin{IEEEproof}
Let $\ha=(\hat{\imath}_0,\ldots,\hat{\imath}_{C})$ be the expert
vector such that the meta experts $(T,a)$ and $(\hT,\ha)$ perform identically.
Then clearly
\begin{eqnarray*}
\lefteqn{\hL_n -  L_n(T,a) }\quad  \\
&= &   \hL_n -  L_n(\A,\hT_n) + L_n(\A,\hT_n) -  L_n(\hT_n,\ha)~.
\end{eqnarray*}
Using \eqref{eq:univbound} and the concavity of $\rho_\E$, we get
\begin{eqnarray}
\lefteqn{L_n(\A,\hT_n) -  L_n(\hT_n,\ha)} \nonumber \\
& =& \!\!\!\sum_{c=0}^{C(\hT_n)}\biggl(
 L_\A(\th_c,\th_{c+1}) - L_{\hat{\imath}_c}(\th_c,\th_{c+1}) \biggr)
  \nonumber \\
&\le &  \!\!\!\sum_{c=0}^{C(\hT_n)} \!\!\rho_\E(\th_{c+1}-\th_c)
\nonumber \\
& \le & 
 (C(\hT_n)+1)\rho_\E\!\!\left(\frac{n}{C(\hT_n)+1}\right). \label{eq:avbound}
\end{eqnarray}

Assume that the loss function $\ell$ is convex in its first
argument and takes values in the interval $[0,1]$. Since
Algorithm~\ref{alg:master} is equivalent to the exponentially weighted
average forecaster with experts $\{(\A,T): T\in \T_n, w_n(T)>0\}$
and initial weights $w_n(T)$, we can apply the bound \eqref{convex2}
to obtain
\[
 \hL_n \le  
L_n(\A,\hT_n)  + \frac{1}{\eta} \ln \frac{1}{w_n(\hT_n)} + \sum_{t=1}^n \frac{\eta_t}{8}.
\]
Combining this with \eqref{eq:avbound} proves \eqref{eq:lemconvex}. 

Now assume $\ell$ is exp-concave.  
Then by \cite[Lemma~1]{HeWa98}, 
\begin{equation}
\label{eq:wbound}
 \hL_n - L_n(\A,\hT_n) 
\le
 \frac{1}{\eta} \ln \frac{1}{w_n(\hT_n)}~.
\end{equation}
This, together with \eqref{eq:avbound}, implies  \eqref{eq:lemexpconcave}.

\end{IEEEproof}

\subsection{The weight function}
\label{subsection:weight}

One may interpret the weight function $\{w_t\}$ as the conditional
probability that a new segment is started, given the beginning of the
current segment and the current time instant. 
In this case one may define $\{w_t\}$ in terms of a 
time-inhomogeneous Markov chain $\{U_t;\, t=1,2,\ldots\}$ whose state
space at time $t$ is $\{1,\ldots,t\}$. Starting from state $U_1=1$, at any time instant $t$, the Markov-chain either stays where it was at
time $t-1$ or switches to state $t$. The distribution of $\{U_t\}$
is uniquely determined by prescribing $\P(U_1=1)=1$ and for $1\le t'<
t$,
\begin{eqnarray}
\label{eq:switch}
\lefteqn{  \P(U_t=t|U_{t-1}=t')} \qquad  \nonumber  \\
&=&   1- \P(U_t=t'|U_{t-1}=t')= p(t|t')
\end{eqnarray}
where the so-called \emph{switch probabilities} $p(t|t')$ need only satisfy
$p(t|t')\in [0,1]$ for all $1\le t'< t$. A realization of this Markov
chain uniquely determines a transition path: $T_t(u_1,\ldots,u_t)=
(t_1,\ldots,t_C)\in \T_t$ if and only if $u_{k-1}\neq u_{k}$ for $k\in
\{t_1,\ldots,t_C\}$, and $u_{k-1}=u_k$ for $k\notin
\{t_1,\ldots,t_C\}$, $2 \le k \le t$. Inverting this correspondence, any $T\in \T_t$
uniquely determines a realization  $(u_1,\ldots,u_t)$.  
Now the  weight function is given for all $t\ge 1$ and $T\in \T_t$ by
\begin{equation}
\label{eq:markov}
  w_t(T) = \P(U_1=u_1,\ldots,U_t=u_t)
\end{equation}
where $(u_1,\ldots,u_t)$ is such that $T=T(u_1,\ldots,u_t)$. It is
easy to check that $\{w_t\}$ satisfies the two conditions in
\eqref{eq:wcond}. Clearly, the switch probabilities $p(t|t')$ uniquely
determine $\{w_t\}$. The above structural assumption on $\{w_t\}$,
originally introduced in \cite{Wil96}, greatly reduces the possible
ways of weighting different transition paths, allowing implementation
of Algorithm~\ref{alg:master} with complexity at most $O(n^2)$ (if one
step of $\A$ can be implemented in constant time), instead of the
potentially exponential time complexity of the algorithm in the naive
implementation; see
Section~\ref{sec_lowcomplex}.

Some examples that have been proposed 
for this construction (given in
terms of the switch probabilities) include

\smallskip

\begin{itemize}

\item $w^{HW}$, used in \cite{HeWa98}, is defined by
$p_{HW}(t|t')=\alpha$ for some $0<\alpha<1$.

\medskip

\item $w^{HS}$, used in \cite{KoRo08, RoEr09, HaSe09}, is defined by
$p^{HS}(t|t')=1/t$.

\medskip

\item $w^{KT}$, used in \cite{Wil96}, is defined by
\begin{equation}
\label{eq:pKT}
p_{KT}(t|t')=\frac{1/2}{t-t'+1}
\end{equation}
which is the Krichevsky-Trofimov  estimate \cite{KrTr81} for binary sequences of the
probability that after observing an all zero sequence of length
$t-t'$, the next symbol will be a one.  Using standard bounds on the
Krichevsky-Trofimov estimate, it is easy to show (see, e.g.,
\cite{Wil96}) that for any $T\in\T_n$ with segment lengths
$s_0,s_1,\ldots,s_C \ge 1$ (satisfying $\sum_{c=0}^C s_c=n$)
\begin{equation}
\label{KTweight}
\ln \frac{1}{w^{KT}(T)} \le \frac{1}{2}\sum_{c=0}^C \ln s_c + (C+1) \ln 2.
\end{equation}

\medskip

\item $w^{\L_1}$ and $w^{\L_2}$ used in \cite{ShMe99} (similar weight functions were considered
in \cite{Vov99}), are defined as follows:
for a given $0<\e<1$,\footnote{The upper bound $\e<1$ is missing from \cite{ShMe99}, although it is implicitly required in the proof.} let $\pi_j=1/j^{1+\e}$,
$Z_t=\sum_{j=1}^t \pi(j)$ (with $Z_0=0$ and $Z_\infty=\sum_{j=1}^\infty \pi(j)$). Then
$w^{\L_1}$  and $w^{\L_2}$ are defined, respectively,  by
\[
p_{\L_1}(t|t')=\frac{\pi(t-1)}{(Z_\infty-Z_{t-2})}
\]
and
\[
p_{\L_2}(t|t')=\frac{\pi(t-t')}{(Z_\infty-Z_{t-t'+1})}.
\]
\end{itemize}

\medskip

Here we consider the weights $w^{\L_1}$.
It is shown in \cite[proof of Eq.\  (39)]{ShMe99} 
that for any $T\in\T_n$,
\begin{equation}
\ln \frac{1}{w^{\L_1}_n(T)}  
\le  
(C_n(T)+\e)\ln n+\ln(1+\e)-C_n(T) \ln \e~. \label{w1bound} 
\end{equation}

\subsection{A low-complexity algorithm}
\label{sec_lowcomplex}

Efficient implementation of Algorithm~\ref{alg:master} hinges on three
factors: (i) Algorithm  $\A$ can be efficiently implemented; (ii)
the exponential weighting step can be efficiently 
implemented; which is facilitated
by (iii) the availability of the losses $L_\A(t',t)$ at each time instant $t$ for all $1\le t' \le t$ in the sense that these losses can be computed efficiently. In what follows we
assume that (i) and (iii) hold and develop a method for (ii) via
constructing a new weight function $\{\hw_t\}$ that significantly
reduces the complexity of implementing Algorithm~~\ref{alg:master}. 

First, we observe that the predictor $\hat{p}_t$ of Algorithm~\ref{alg:master}
can be rewritten as
\begin{equation}
\label{eq:implementPred}
\wh{p}_t=\frac{\sum_{t'=1}^t v_t(t') f_{\A,t}(t')}
{\sum_{t'=1}^t v_t(t')}
\end{equation}
where the weights $v_t$ are given by
\begin{equation}
\label{eq:vtdef}
v_t(t')= \sum_{T \in \T_t:\, \tau_t(T)=t'} w_t(T) e^{-\eta_t L_{t-1}(\A,T_{t-1})}.
\end{equation}
Note that $v_t(t')$ gives the weighted sum of the exponential weights of
all transition paths with the last switch at $t'$.

If the learning parameters $\eta_t$ are constant during the time
horizon, the above
means that Algorithm~\ref{alg:master} can be implemented efficiently
by keeping a weight $v_t(t')$ at each time instant $t$ for every
possible starting point of a segment $t'=1,\ldots,t$. Indeed, if
$\eta_t=  \eta$ for all $t$, then \eqref{eq:vtdef},
\eqref{eq:switch}, and \eqref{eq:markov} imply that each $v_t(t')$ can
be computed recursively in $O(t)$ time from the $v_{t-1}$ (setting $v_1(1):=1$
at the beginning) using the switch probabilities defining $w_t$ as
follows:
\begin{equation}
\label{eq:implementWeight}
v_t(t')=
\begin{cases}
v_{t-1}(t')(1-p(t|t'))e^{-\eta\ell(f_{\A,t-1}(t'),y_{t-1})} & \\
\quad\quad\quad \quad\quad\quad \quad\quad\quad\quad  \text{for  $t'= 1,\ldots,t-1$,} & \\
\sum_{t''=1}^{t-1}
v_{t-1}(t'')p(t|t'')e^{-\eta\ell(f_{\A,t-1}(t''),y_{t-1})} & \\
\quad\quad\quad \quad\quad\quad \quad\quad\quad\quad  \text{for $t'=t$.} &
\end{cases}
\end{equation}
Using this recursion, the overall complexity of computing the weights
during $n$ rounds is $O(n^2)$. Furthermore, \eqref{eq:implementPred}
means that one needs to start an instance of $\A$ for each
possible starting point of a segment.  If the complexity of running
algorithm $\A$ for $n$ time steps is $O(n)$ (i.e., computing
$\A$ at each time instant has complexity $O(1)$), then the overall
complexity of our algorithm becomes $O(n^2)$.

It is clearly not a desirable feature that the amount of computation
per time round grows (linearly) with the horizon $n$. 
While we don't know how to completely eliminate this ever-growing computational
demand, we are able to moderate this growth significantly. 
To this end, we modify the weight functions in such a way that at any time instant $t$ we allow at most $O(\g \ln t)$ 
actual segments with positive probability (i.e., segments containing $t$ that belong to sample paths with positive weights),
where $\g>0$ is a parameter of the algorithm (note that $\g$ may
depend on, e.g., the time horizon $n$). Specifically, 
we will construct a new weight function $\hat{w}_t$ such that 
\[
\bigl|\{\tau_t(T): \hat{w}_t(T_t)>0, T\in\T_n\}\bigr|\le \left\lceil \frac{g}{2} \right\rceil (\lfloor \log t \rfloor + 1)
\]
where $\log$ denotes base-2 logarithm.
By doing so, the time and space complexity of the algorithm becomes
$O(\g \ln n)$ times more than that of algorithm $\A$, as
we need to run $O(\g \ln n)$ instances of $\A$ in parallel and the
number of non-zero terms in \eqref{eq:implementWeight} and
\eqref{eq:implementPred} is also $O(\g \ln n)$ (here we exclude the trivial case where $\A$ has zero space
complexity; also note that the time-complexity of $\A$ is at least linear
in $n$ since it has to make a prediction at each time instant).
Thus, in case of a
linear-time-complexity algorithm $\A$, the overall complexity of
Algorithm~\ref{alg:master} becomes $O(\g n\ln n)$.

In order to construct the new weight function, at each time instant
$t$ we force some segments to end. Then any path that contains such a
segment will start a new segment at time $t$ (and hence the
corresponding vector of transitions contains $t$). Specifically, any
time instant $s$ can be uniquely written as $o 2^u$ with $o$ being a
positive odd number and $u$ a nonnegative integer (i.e., $2^u$ is the
largest power of $2$ that divides $s$). We specify that a segment starting
at $s$ can  ``live'' for at most $\g 2^u$ time
instants, where $\g>0$ is a parameter of the algorithm, so that at
time $s+\g 2^u$ we force a switch in the path. More precisely, given
any switch probability $p(t|t')$ for all $t'<t$, we define a new
switch probability
\begin{equation}
\label{eq:phat}
\hp(t|t')= 1- h_t(t')\bigl(1-p(t|t')\bigr)
\end{equation}
where 
\[
h_t(s)=\begin{cases} 1& \text{if $s\le t < s+\g 2^u$,} \\ 
$0$ & \text{otherwise.}
\end{cases}
\]
Thus $h_t(s)=1$ if and only if a segment started at $s$ is still valid
at time $t$. 
In terms of the Markov chain $\{U_t\}$ introduced in \eqref{eq:switch}, the new switch
probabilities in definition \eqref{eq:phat} mean that if the chain is in state
$t'$ at time $t-1$ such that $h_t(t')=1$, then the chain switches to
state $t$ with the original switch probability $p(t|t')$ and remains
at state $t'$ with probability $1-p(t|t')$; but if $h_t(t')=0$, then
the chain switches to state $t$ with probability 1.
In this way, given the switch probabilities  $p(t|t')$  and the
associated weight function $\{w_t\}$, we can define a new weight function
$\{\hw_t\}$ via the new  switch 
probabilities $\hat{p}(t|t')$ and the procedure described in
Section~\ref{subsection:weight}.
Note that the definition of $\{\hat{w}_t\}$ implies that for a transition path $T \in \T_t$ either 
\begin{equation}
\label{eq:w-hw}
\hw_t(T)=0 \quad \text{ or } \quad \hw_t(T) \ge w_t(T)~.
\end{equation}

The above procedure is a common generalization of several algorithms previously reported
in the literature for pruning the transition paths. Specifically, $\g=1$ yields the procedure
of \cite{WiKr97}, $\g=3$ yields our previous procedure
\cite{GyLiLu08b}, $\g=4$ yields the method of \cite{HaSe09}, while
$\g=n$ yields the original weighting $\{w_t\}$ without pruning.  We
will show that the time complexity of the method with a constant $\g$
(i.e., when $\g$ is independent of the time horizon $n$)
is, in each time instant, at most $O(\ln n)$ times the complexity of
one step of $\A$, while the time complexity of the algorithm without
pruning is $O(n)$ times the complexity of $\A$. Complexities that
interpolate between these two extremes can be achieved by setting $\g=o(n)$ appropriately.

We say that a segment at time instant $t$ is \emph{alive} if it
contains $t$ and is \emph{valid} if there is a path $T_t$ with
$\hw_t(T_t)>0$ that contains exactly that segment. In what follows we
assume that the original switch probabilities $p(t|t')$ associated
with the $w_t$ satisfy $p(t|t')\in (0,1)$ for all $1\le t'< t$. (Note
that the weight function examples introduced in
Section~\ref{subsection:weight} all satisfy this condition.) The condition
implies that $w_t(T_t)>0$ for all $T_t \in \T_t$. Furthermore, if
$T_t=(t_1,\ldots,t_C)\in \T_t$ satisfies $t_{i+1}-t_i <
\g 2^{u_{t_i}}$, $i=1,\ldots,C$, where $u_{t_i}$ is the largest
power of 2 divisor of $t_i$, then from \eqref{eq:phat} we get
$\hw_t(T)>0$.

The next lemma gives a characterization of when $h_t(s)=1$ and, as a
consequence, bounds the number of valid segments that are alive at
$t$. 

\medskip 

\begin{lemma}
\label{lem:h_t(s)}
Let $t=\sum_{i=1}^m 2^{u_i}$ be the binary form of $t$ with
$0 \le u_1<u_2<\cdots<u_m$, $s_k=\sum_{i=k}^m 2^{u_i}$, and $u_0=-1$.
Then $h_t(s)=1$ if and only if $s=s_k-j 2^u$ for some $u_{k-1} < u \le u_k$ and $j\in\{0,\ldots,g-1\}$ such that
$2^u$ is the largest $2$-power divisor of $s$; in particular, $j$ is
even if $u=u_k$ for some $k\in \{1,\ldots,m\}$, and odd otherwise.
As a consequence, at any time instant $t$ there are at most $\lceil \g/2 \rceil (\lfloor \log t \rfloor + 1)$ segments that are valid and alive.
\end{lemma}

\medskip 

\begin{IEEEproof}
It is clear that for any $s$ satisfying the conditions of the lemma, $h_t(s)=1$ since $s+\g 2^u = s_k-j 2^u + \g 2^u \ge
s_k + 2^u >t \ge s$. To prove the other direction, consider an $s \in
[1,t]$; assume $h_t(s)=1$ and denote the largest $2$-power divisor of
$s$ by $2^u$. 
By definition, $h_t(s)=1$ if and only if $s+j2^u \le t < s+(j+1) 2^u$
for some $j\in \{0,\ldots,\g-1\}$. After reordering we obtain 
\begin{equation}
\label{eq:s-tj}
t-(j+1) 2^u < s \le t-j 2^u.
\end{equation}
Let $k \in \{1,\ldots,m\}$ be the unique index such that $u_{k-1}<u \le
u_k$ (note that $u\le u_m$ always holds). Then $2^u$ divides $s_k$, and
$s_k \le t < s_k+2^u$. Combining this inequality  with \eqref{eq:s-tj} gives
$s_k-(j+1) 2^u < s < s_k-(j-1) 2^u$. Taking into account that both $s$
and $s_k$ are divisible by $2^u$, we
obtain $s=s_k-j 2^u$. Furthermore, since $2^u$ is the largest
$2$-power divisor of $s$, $j$ must be even when $u=u_k$ for some
$k\in \{1,\ldots,m\}$, and odd otherwise.

Finally, for any $u\in \{0,1,\ldots,u_m\}$, the set 
\begin{eqnarray*}
\lefteqn{\bigl\{s=s_k-j 2^u: u_{k-1} < u \le u_k, j=0,\ldots,\g-1,} \\
&& \quad\quad \quad\quad \quad\quad\text{$2^u$ is the largest $2$-power divisor of $s$} \bigr\}
\end{eqnarray*}
has at most $\lceil \g/2 \rceil$ elements. Since $u_m=\lfloor \log t
\rfloor$, the proof is complete.
\end{IEEEproof}

Note that for $\g= 1$ the valid segments that are alive at $t$ start exactly at $s_k, k=1,\ldots,m$, and so the number of valid segments at time $t$
is exactly the number of $1$'s in the binary form of $t$ \cite{WiKr97}.
The above lemma implies that Algorithm~\ref{alg:master} can be
implemented efficiently with the proposed weight function $\{\hw_t\}$.

\medskip

\begin{theorem}
\label{thm:complexity}
Assume Algorithm~\ref{alg:master} is run with weight function
$\{\hw_t\}$ derived using any $g>0$ from any weight function
$\{w_t\}$  
defined as in Section~\ref{subsection:weight}.
If $\eta_t= \eta$ for some $\eta>0$ and all $t=1,\ldots,n$, then
the time and space complexity of  
Algorithm~\ref{alg:master}  is $O(\g \ln n)$ times the time and space complexity of $\A$, respectively.
\end{theorem}

\medskip 

\begin{IEEEproof}
The result follows since  Lemma~\ref{lem:h_t(s)} implies that the
number of   non-zero terms in \eqref{eq:implementWeight}
and \eqref{eq:implementPred} is always  $O(\g \ln t)$. 
\end{IEEEproof}

\subsection{Regret bounds}
\label{subsection:regret}

To bound the regret, we need the following lemma which shows that any
segment $[t,t')$ can be covered with at most $\left\lceil \frac{\log
    (t'-t)}{\lfloor \log (\g+1) \rfloor} \right\rceil + 1$ valid
  segments.

\medskip 

\begin{lemma}
\label{lem:segmentlength}
For any $T\in T_n$, there exists  $\hT \in T_n$ such that for any segment $[t,t')$ of $T$ with $1\le t<t'\le n+1$, 
\begin{enumerate}[(i)]
\item  $\hw_{t'}(\hT)>0$,  $t$ and $t'$ are switch points of
  $\hT$ (if $t'=n+1$, it is considered as a switch point), and 
$\hT$ contains at most $l=\left\lceil \frac{\log(t'-t)}{\lfloor \log (\g+1) \rfloor} \right\rceil + 1$ segments in  $[t,t')$;
\item if the switch points of $\hT$ in $[t,t')$ are
  $t_1:=t<t_2<\cdots<t_{l'}<t_{l'+1}:=t'$, then $l' \le l$, and
  for any nondecreasing function $f:[0,\infty)\to [0,\infty)$,
\begin{eqnarray}
\label{eq:segmentlength-refined}
\lefteqn{\!\!\!\! \!\!\!\!  \sum_{i=1}^{l'} f(t_{i+1}-t_i) } \nonumber \\
& \!\!\!\! \!\!\!\! \!\!\!\!\!\!\!\!   \le & \!\!\!\!   \!\!\!\! \!\!\sum_{i=0}^{l'-2} f\left( \frac{t'-t}{2^{i
    \left\lfloor \log(\g+1) \right\rfloor}}
\right)+f (t'-t) \\*
&\!\!\!\! \!\!\!\! \!\!\!\! \!\!\!\!   \le& \!\!\!\!   \!\!\!\! \!\!\int_0^{\frac{\log(t'-t)}{\lfloor \log (\g+1) \rfloor}}\! f\!\left(\frac{t'-t}{2^{x \left\lfloor \log(\g+1) \right\rfloor}} \right) dx 
+2 f(t'-t)  \label{eq:segmentlength-refined2}
\end{eqnarray}
where the second summation in \eqref{eq:segmentlength-refined} is
empty if $l'=1$. 
\end{enumerate}

\end{lemma}

\emph{Remark:} \ 
Note that it is possible to obtain for $l$ the less compact and harder-to-handle formula
\[
l=
\left\lceil \frac{\log\frac{t'-t+\frac{1}{2^{\lfloor \log (\g+1) \rfloor}-1}}{2^{\lfloor \log (\g+1) \rfloor}-1+\frac{1}{2^{\lfloor \log (\g+1) \rfloor} -1}}}{\lfloor \log (\g+1) \rfloor}\right\rceil +1
\]
by taking into account that the last segment $[t_l,t_{l+1})$ in the construction of the proof can always be defined to be of length 
at least  $\lfloor \log (\g+1) \rfloor 2^{u_l}$. Furthermore,
for $\g= 1$ it follows from \cite{WiKr97} that the last term is not needed in (\ref{eq:segmentlength-refined}), and hence the latter bound  can be strengthened to
\begin{equation}
\label{eq:segmentlength-refined1}
\sum_{i=1}^{l'} f(t_{i+1}-t_i) \le \sum_{i=0}^{\lfloor \log(t'-t) \rfloor} f(2^i).
\end{equation}

\medskip 

\begin{IEEEproof}
Clearly, it is enough to define $\hT$ independently in each segment $[t,t')$ of $T$. We construct the switch points $t_1<t_2<\cdots<t_{l'}$ of $\hT$ in this interval, for some $l'\le l$, and an auxiliary variable $t_{l'+1} \ge t'$ one by one such that
$t_1=t$, $t_{l'}<t'$ and, defining $u_j$ as the largest $2$-power divisor of $t_j$,
\begin{equation}
\label{eq:u-increment}
u_{j+1}-u_j \ge \lfloor  \log(\g+1) \rfloor
\end{equation} 
for $j=1,\ldots,l'-1$. Assume that we have already defined
$t_1,\ldots,t_i$ satisfying \eqref{eq:u-increment} for
$j=1,\ldots,i-1$. Then a segment starting at $t_i$ may be alive with
positive probability at any time instant in $[t_i,t_i+\g
  2^{u_i})$. Define $u_{i+1}$ to be the largest nonnegative integer
  such that there is an $s \in [t_i+1, t_i+\g 2^{u_i}]$ such that
  $2^{u_{i+1}}$ divides $s$. Then $s2^{-u_i}$ belongs to the set
  $\mathcal{S}_i=\{t_i 2^{-u_i}, t_i 2^{-u_i}+1, t_i
  2^{-u_i}+2,\ldots,t_i 2^{-u_i}+\g\}$ (although, clearly,
  $s2^{-u_i}\neq t_i 2^{-u_i}$). Since $\mathcal{S}_i$ is a set of
  $\g+1$ consecutive integers, it has an element $a$ that is divisible by
  $2^{\lfloor \log(\g+1) \rfloor}$, and this element is not the odd
  number $t_i 2^{-u_i}$. Thus $2^{u_i}a \in [t_i+1,t_i+g2^{u_i}]$ and
  since $2^{u_i}a$ is divisible by $2^{u_i+\lfloor \log
    (g+1)\rfloor}$, the maximal property of the $2$-power divisor
  $2^{u_{i+1}}$ of $s$ implies that  $u_{i+1} \ge u_i+\lfloor \log
  (\g+1) \rfloor$. Therefore, defining 
  $t_{i+1}=s$, its largest $2$-power divisor is $2^{u_{i+1}}$, proving
  \eqref{eq:u-increment} for $j=i$ (note that it is easy to show that
  the choice of $s$, and hence that of $t_{i+1}$, is unique).

Now let $l'$ be the smallest integer such that $t_{l'+1} \ge t'$. To
prove part (i) of the lemma, it is sufficient to show that $l' \le l$
and the segments
$[t_1,t_2),[t_2,t_3),\ldots,[t_{l'-1},t_{l'}),[t_{l'},t')$ cover
        $[t,t')$, which is clearly true if $t_{l+1} \ge t'$. 
From \eqref{eq:u-increment} and the fact that $t_{i+1}-t_i$ is
divisible by $2^{u_i}$, we have
\begin{eqnarray*}
t_{l+1} &\ge& t+ \sum_{i=1}^l 2^{u_i} = t+ \sum_{i=1}^l 2^{u_1+\sum_{j=2}^i (u_j-u_{j-1})} \\
&\ge& t+ \sum_{i=1}^l 2^{u_1+\sum_{j=2}^i \lfloor \log(\g+1) \rfloor} \\
&=&  t+ \sum_{i=0}^{l-1} 2^{u_1+i \lfloor \log(\g+1) \rfloor} \\*
&=& t+2^{u_1}\frac{2^{l \lfloor \log(\g+1) \rfloor}-1}{2^{\lfloor\log(\g+1) \rfloor}-1} \\ 
&\ge& t + 2^{(l-1) \lfloor\log(\g+1) \rfloor}  
\ge t'
\end{eqnarray*}
where in the last step we used the definition of $l$. This finishes the proof of (i).

To prove (ii), we first show that the transition path $\hT$
constructed above satisfies \eqref{eq:segmentlength-refined}, where,
with a slight abuse of notation, we redefine $t_{l'+1}$ from part (i)
to be $t'$. First notice that since $t+\g 2^{u_{l'-1}} \le t_{l'-1}+\g
2^{u_{l'-1}} < t'$, we have $u_{l'-1} \le \left\lfloor
\log\frac{t'-t}{\g} \right\rfloor $. Repeated application of
\eqref{eq:u-increment} implies, for any $i=1,\ldots,l'-1$,
\[
u_{i} \le \left\lfloor \log\frac{t'-t}{g} \right\rfloor- (l'-1-i) \left\lfloor \log(\g+1) \right\rfloor
\]
and
\begin{eqnarray*}
t_{i+1}-t_i &\le & \g 2^{\left\lfloor \log\frac{t'-t}{g}
  \right\rfloor- (l'-1-i) \left\lfloor \log(\g+1) \right\rfloor} \\*
&\le & \g 2^{\log\frac{t'-t}{g} - (l'-1-i) \left\lfloor \log(\g+1) \right\rfloor} \\
&=& (t'-t) 2^{- (l'-1-i) \left\lfloor \log(\g+1)
  \right\rfloor}.
\end{eqnarray*}
Using the crude estimate $t'-t_l \le t'-t$ finishes the proof of \eqref{eq:segmentlength-refined}.
The last inequality \eqref{eq:segmentlength-refined2} holds trivially
for $l=1$, and holds for $l \ge 2$ since 
\begin{eqnarray*}
\lefteqn{\sum_{i=0}^{l'-2} f\left( \frac{t'-t}{2^{i \left\lfloor \log(\g+1) \right\rfloor}} \right)} \\
&=& f (t'-t) +\sum_{i=1}^{l'-2} f\left( 
\frac{t'-t}{2^{i  \left\lfloor \log(\g+1) \right\rfloor}} \right) \\*
&\le& f (t'-t) +\int_{0}^{\left\lceil
  \frac{\log (t'-t)}{\lfloor \log (\g+1) \rfloor} \right\rceil-1}
f\left( \frac{t'-t}{2^{x \left\lfloor
    \log(\g+1) \right\rfloor}} \right) \, dx\\ 
&\le& f(t'-t) +\int_0^{\frac{\log(t'-t)}{\lfloor \log (\g+1) \rfloor}}
f\left( \frac{t'-t}{2^{x \left\lfloor \log(\g+1) \right\rfloor}} \right) \, dx. 
\end{eqnarray*} 
\end{IEEEproof}

\medskip

Taking into account that $C(T_n) \le C(\hT_n)$ if $\hT_n$ covers $T_n$, 
Lemma~\ref{lem:segmentlength} trivially implies the following bounds.

\medskip

\begin{lemma}
\label{lem:Cbound}
For any $T_n
\in \T_n$ there 
exists a $\hT_n\in\T_n$ with $\hw_n(\hT_n)>0$ such that $\hT_n$ covers
$T_n$ and 
\begin{equation}
\label{eq:Cbound}
C(T_n) \le C(\hT_n) \le (C(T_n)+1) L_{C(T_n),n} -1 
\end{equation}
where  
\begin{equation}
\label{eq:L_cn}
L_{C,n}=\begin{cases}
\left \lceil  \frac{\log n}{\lfloor \log (\g+1)\rfloor} \right\rceil +1 & \text{if $C=0$,} \\
\frac{\log \frac{n}{C+1}}{\lfloor \log (\g+1) \rfloor} +2 & \text {if
  $C \ge 1$}. 
\end{cases}
\end{equation}
\end{lemma}

\medskip

\begin{IEEEproof}
The lower bound is trivial, and the upper bound directly follows from Lemma~\ref{lem:segmentlength} for $C(T_n)=0$.
For $C(T_n) \ge 1$ the upper bounds follow since on each segment of $T_n$ we can define $\hT_n$ as in the proof of
Lemma~\ref{lem:segmentlength}. Hence, if $T=(t_1,\ldots,t_C;n)$, then
\begin{eqnarray*}
C(\hT_n)+1 &\le& \sum_{i=1}^{C+1}\left(\left \lceil \frac{\log(t_i-t_{i-1})}{\lfloor \log(\g+1) \rfloor} \right\rceil +1\right) \\
&\le& \sum_{i=1}^{C+1}\left( \frac{\log(t_i-t_{i-1})}{\lfloor \log(\g+1) \rfloor} +2 \right) \\*
&\le& (C+1)\left(\frac{\log\frac{n}{C+1}}{\lfloor \log(\g+1) \rfloor} +2\right)
\end{eqnarray*}
where in the last step we used Jensen's inequality and the concavity
of the logarithm.  
\end{IEEEproof}

\medskip

We now apply the preceding construction  and results to the weight
function $\{w_t\}= \{w^{\L_1}_t\}$ to obtain our main theorem:

\medskip

\begin{theorem}
\label{thm:main}
Assume Algorithm~\ref{alg:master} is run with $\g >0$  and weight function
 $\{\hw^{\L_1}_t\}$ for some $0<\e<1$ (derived from  $\{w^{\L_1}_t\}$), based on a prediction algorithm that satisfies
(\ref{eq:univbound}) for some $\rho_\E$.
Let $L_{C,n}$ be defined by \eqref{eq:L_cn}.
If $\ell$ is convex in its first argument and takes values in the interval $[0,1]$
and $\eta_{t+1} \le \eta_t$ for $t=1,\ldots,n-1$,
then for all $n \ge 1$ and any $T \in \T_n$,
the  tracking regret satisfies
\begin{eqnarray}
\lefteqn{\hL_n -  L_n(T,a)} \nonumber \\
&\le &
L_{C(T),n} (C(T)+1) \rho_\E\left( \frac{n}{L_{C(T),n} (C(T)+1)}\right)
 \nonumber \\*
&  & + \mbox{} \sum_{t=1}^n \frac{\eta_t}{8} + \frac{r_n\left( L_{C(T),n} (C(T)+1)-1\right)}{\eta_n}
\label{eq:thmconvex}
\end{eqnarray}
where the function $r_n(C)$ is defined as
\[
r_n(C)= (C+\e)\ln n+\ln(1+\e)-C \ln \e.
\]
Furthermore, for $\e \le 1/2$ and $n \ge 5$, the adaptive regret of the algorithm satisfies 
\begin{equation}
\label{eq:adaptive1}
R_n^a \le  
L_{0,n} \rho_\E\left( \frac{n}{L_{0,n}}\right) +\sum_{t=1}^n \frac{\eta_t}{8} + \frac{r'_n\left(L_{0,n}-1\right)}{\eta_n}
\end{equation}
where the function $r'_n(C)$ is defined as
\[
r'_n(C)=(C+1)\ln n -(C+1) \ln \e.
\]
On the other hand, if $\ell$ is exp-concave for some $\eta>0$ and we let
$\eta_t = \eta$ for $t=1,\ldots,n$ in Algorithm~\ref{alg:master}, then
for any $n \ge 1$ and $T \in \T_n$ 
the tracking regret satisfies
\begin{eqnarray}
\lefteqn{\hL_n -  L_n(T,a)} \nonumber \\ 
& \le & 
L_{C(T),n} (C(T)+1) \rho_\E\left( \frac{n}{L_{C(T),n} (C(T)+1)}\right)
\nonumber \\*
 & & \mbox{} + \frac{r_n\left(L_{C(T),n}
  (C(T)+1)-1\right)}{\eta} \label{eq:thmexpconcave} 
\end{eqnarray}
while for $0<\e \le 1/2$ and $n \ge 5$, the adaptive regret can be bounded as
\begin{equation}
\label{eq:adaptive2}
R_n^a \le  
L_{0,n} \rho_\E\left( \frac{n}{L_{0,n}}\right) + \frac{r'_n\left(L_{0,n}-1\right)}{\eta}~.
\end{equation}
\end{theorem}
\begin{IEEEproof}
First we show the bounds for the tracking regret.
To prove the theorem, let $\hT_n$ be defined as in Lemma~\ref{lem:master},  and we bound the first and last terms on the right-hand side
of \eqref{eq:lemconvex} and \eqref{eq:lemexpconcave} (with $\hw^{\L_1}_n$ in
place of $w_n$). 
Note that the conditions on $\rho_\E$ imply that
$x \rho_\E(y/x)$ is a nondecreasing function of $x$ for any
fixed $y>0$  (this follows since $\rho_\E(z)/z = (\rho_\E(z)-0)/(z-0)$
is a nonincreasing function of $z>0$ by the concavity of $\rho_\E$,
and hence $z\rho_\E(1/z)$ is nondecreasing).
Combining this  with the bounds
on $C(T_n)$ in Lemma~\ref{lem:Cbound} implies
\begin{eqnarray*}
\lefteqn{(C(\hT_n)+1)\rho_\E\left(\frac{n}{C(\hT_n)+1}\right)} \\ 
&\le& L_{C(T),n} (C(T)+1) \rho_\E\left( \frac{n}{L_{C(T),n} (C(T)+1)}\right).
\end{eqnarray*}
The last term $(1/\eta_n)\ln(1/\hw_n^{\L_1}(\hT_n))$ in \eqref{eq:lemconvex} and \eqref{eq:lemexpconcave} can be bounded
by noting that $1/\hw_n^{\L_1}(\hT_n) \le 1/w_n^{\L_1}(\hT_n)$ by \eqref{eq:w-hw} and the latter can be bounded using \eqref{w1bound};
this is given by $r_n$.
This finishes the proof of the tracking regret bounds.

Next we prove the bounds for the adaptive regret. Assume we want to
bound the regret of our algorithm in a segment $[t,t')$ with $1\le
  t<t' \le n+1$. By Lemma~\ref{lem:segmentlength} there exists a
  transition path $\hT_{t'-1}$ such that it has a switch point at
  $t$, has at most $l=\left\lceil \frac{\log(t'-t)}{\lfloor \log(g+1)
    \rfloor}\right\rceil+1 \le L_{0,n}$ segments in $[t,t')$, and
    $\hw_n(\hT_n)>0$. Let $\th_1,\th_2,\ldots,\th_{\hC}$ denote the
    switch points of $\hT_n$ in $[t+1,t')$ where $\hC < l$, and let
      $\th_0=t$ and $\th_{\hC+1}=t'$. Notice that, since we are
      interested in the performance of the algorithm only in the
      interval $[t,t')$, a modified version of Lemma~\ref{lem:master}
        can be applied, where the loss is considered only in the
        interval $[t,t')$ and the weight of $\hT_n$ can be thought to
          be the sum of the weight of all transition paths that agree
      with $\hT_n$ in $[t,t')$. Specifically, letting
            $\T_{t,t'}(\hT_{t'-1})=\{T\in\T_{t'-1}: \text{$T$ and
              $\hT_{t'-1}$ agree on $[t,t')$}\}$ and
              $\hw^{\L_1}_{t,t'}(\hT_n)=\sum_{T \in \T_{t,t'}}
              \hw^{\L_1}_{t'-1}(T)$, it can be shown similarly to
              Lemma~\ref{lem:master} that in the case of a  loss
              function that is convex in its first argument and takes
              values in $[0,1]$,   for any expert
              $i\in\E$, 
\begin{eqnarray}
\lefteqn{\sum_{s=t}^{t'-1} \left(\ell(\hp_s,y_s) - \ell(a,y_s)\right)}
\nonumber\qquad  \\
&\le & (\hC+1)\rho_\E\!\!\left(\frac{n}{\hC+1}\right) \nonumber \\*
 & & \mbox{} + \sum_{s=t}^{t'-1} \frac{\eta_s}{8} + \frac{1}{\eta_{t'-1}} \ln \frac{1}{\hw_{t,t'}(\hT_{t'-1})}~.
\label{eq:convex-mod}
\end{eqnarray}
Now $-\ln \hw^{\L_1}_{t,t'}(\hT_{t'-1})$ can be bounded in a similar way as $-\ln \hw^{\L_1}_n(T_n)$ in \cite{ShMe99}: For $t=1$ we can use \eqref{w1bound}. For $t \ge 2$ it can be shown, following the proof of \eqref{w1bound} in \cite{ShMe99}, that 
\begin{eqnarray}
\ln \frac{1}{\hw^{\L_1}_{t,t'}(\hT_{t'-1})} 
&\le& (\hC+1) \ln (t'-1) - (\hC+1) \ln \e \nonumber \\
&\le&  (\hC+1) \ln n - (\hC+1) \ln \e
\label{w1bound-mod}
\end{eqnarray}
whenever $\e \le 1/2$. 
Indeed, let $B_t$ denote the event that $t$ is a switch point and let
$A_{t_1,\ldots,t_{\hC}}$ denote the event that $t_1,\ldots,t_{\hC}$ are
the switch points in $[t+1,t')$. Since the switch probabilities $p_{\L_1}(s|s')$ are independent of $s'$ and 
$1-p_{\L_1}(s|s')=\frac{Z_\infty-Z_{s-1}}{Z_\infty-Z_{s-2}}$,
for $\e \le 1/2$, we have 
\begin{eqnarray*}
\lefteqn{\hw^{\L_1}_{t,t'}(\hT_{t'-1}) } \\
&=& \P(B_t)\P(A_{t_1,\ldots,t_{\hC}} |B_t) \\*
&\ge& \prod_{c=0}^{\hC} \frac{\pi(t_c-1)}{Z_\infty-Z_{t_c-2}} 
\biggl( \;\prod_{\tau=t_c+1}^{t_{c+1}-1}
\frac{Z_\infty-Z_{\tau-1}}{Z_\infty-Z_{\tau-2}}\biggr)  \\
&=& \prod_{s=t}^{t'-1} \frac{Z_\infty-Z_{s-1}}{Z_\infty-Z_{s-2}} \cdot \prod_{c=0}^{\hC} \frac{\pi(t_c-1)}{Z_\infty-Z_{t_{c-1}}} \\
&=& \frac{Z_\infty-Z_{t'-2}}{Z_\infty-Z_{t-2}} \prod_{c=0}^{\hC} \frac{\pi(t_c-1)}{Z_\infty-Z_{t_{c-1}}}  \\
& \ge & \frac{(t-1)^{1+\e}}{(t'-1)^\e (t-1+\e)}\cdot \frac{\e t^{1+\e}}{(t+\e)(t-1)^{1+\e}}\cdot\frac{\e^{\hC}}{(t'-1)^{\hC}} \\
& = & \frac{\e^{\hC+1} t^{1+\e}}{(t'-1)^{\hC+\e}(t-1+\e)(t+\e)}  \\
& \ge&   \frac{\e^{\hC+1}}{(t'-1)^{\hC+\e}\ t^{1-\e}} \ge \frac{\e^{\hC+1}}{(t'-1)^{\hC+1}}\\
\end{eqnarray*}
where the second  inequality follows form inequalities (36) and (38) in
\cite{ShMe99}, and the third  follows since $(t-1+\e)(t+\e)<t^2$.

It is easy to see that the bound in \eqref{w1bound-mod} is larger than \eqref{w1bound} if $n \ge 5$. Thus, combining
with \eqref{eq:convex-mod} for the maximizing value $t=1$, $t'=n+1$ and using $\hC \le L_{0,n}$, we obtain the bound \eqref{eq:adaptive1} on the adaptive regret. A modified version of \eqref{eq:convex-mod} (without the $\sum_{s=t}^{t'-1} \eta_s/8$ term) yields \eqref{eq:adaptive2}
\end{IEEEproof}

\medskip

\emph{Remarks:} \ 
(i) Note that the tracking regret can be trivially bounded by
$(C(T)+1)$ times the adaptive regret (as suggested by
\cite{HaSe09}). However, the direct bounds on the tracking regret are
somewhat better than this: The first term coming from the adaptive
regret bound would be $L_{0,n}(C(T)+1)\rho_\E(n/L_{0,n})$, which is
larger than the first term
$L_{C,n}(C(T)+1)\rho_\E(\frac{n}{L_{C,n}(C(T)+1)})$ in the tracking
regret bounds. This justifies our claim for exp-concave loss
functions, since the last terms will be essentially the same, although
the main term in the bound is not affected. The difference is more
pronounced for the case of the convex and bounded loss function, where
the middle $\sum_t \eta_t/8$ term becomes multiplied by $(C(T)+1)$ if
the tracking bound is computed from the adaptive regret bound,
resulting in an increased constant factor in the main
term. \\
(ii) The above theorem provides bounds on the tracking and adaptive regrets in
terms of the regret bound $\rho_\E$ of algorithm $\A$. However, in
many practical situations, $\A$ behaves much better than suggested by
its regret bound. This behavior is also preserved in our tracking
algorithms: Omitting step \eqref{eq:avbound} in Lemma~\ref{lem:master}
we can replace the first term in \eqref{eq:thmconvex} and
\eqref{eq:thmexpconcave} with $L_n(\A,\hT_n) - L_n(\hT_n,\ha)$, which
is the actual regret of algorithm $\A$ on the (extended) transition
path $\hT_n$. Reordering the resulting inequality, we can see that the
loss of our algorithm is not much larger than that of $\A$ run on
$\hT_n$, for example, in the exp-concave case we have
\[
\hL_n - L_n(\A,\hT_n) \le  \frac{r_n\left(L_{C(T),n} (C(T)+1)-1\right)}{\eta}.
\]

\subsection{Exponential weighting}
\label{subsection:exp}
We now apply Theorem~\ref{thm:main} to the case where $\A$ is the
exponentially weighted average forecaster and the set of base experts
is of size $N$, and discuss the obtained  bounds (for simplicity we
assume $C(T) \ge 1$, but $C(T)=0$ would just slightly change the
presented bounds).   In this case, if $\ell$ is convex and bounded,
then  by \eqref{convex} the regret of  $\A$ is bounded by
$\rho_{\E}(n)= 
\sqrt{n\ln N}$. Setting
$\eta_t\equiv\phi \ln n/\sqrt{n}$ for some $\phi>0$ ($\eta_t$ is independent of $C(T)$ but depends on the time horizon $n$), the bound
\eqref{eq:thmconvex} becomes, for  $\g=O(1)$,
\begin{eqnarray*}
\lefteqn{\hL_n -  L_n(T,a) } \\
& \le &\sqrt{n (C(T)+1) \left(\frac{\log n}{\lfloor  \log(\g+1) \rfloor}+2\right) \ln N} \\*
& & \mbox{}+\frac{\phi \sqrt{n}\ln n}{8}
+\frac{(C(T)+1)\sqrt{n}}{\phi}\left(\frac{\log n}{\lfloor  \log(\g+1) \rfloor}+2\right) \\*
& & \mbox{}+ O\left(\frac{\sqrt{n}}{\ln n}\right)~.
\end{eqnarray*}
Furthermore, if an upper bound $C$ on the complexity (number of
switches) of the meta experts in the reference class is known in
advance, then $\eta_t$ can be set as a function of $C \ge C(T)$ as
well. Letting $\eta_t\equiv\sqrt{8(C+1) \ln n \left(\frac{\log
    n}{\lfloor  \log(\g+1) \rfloor}+2\right)/n}$,  the bound \eqref{eq:thmconvex} becomes
\begin{eqnarray*}
\lefteqn{\hL_n -  L_n(T,a) }\\
& \le &\sqrt{n (C(T)+1) \left(\frac{\log n}{\lfloor  \log(\g+1) \rfloor}+2\right) \ln N} \\*
& & \mbox{} +
\sqrt{\frac{n(C+1) \left(\frac{\log n}{\lfloor  \log(\g+1) \rfloor }+2\right) \ln n}{2}} \\*
& & \mbox{} + O\left(\sqrt{\frac{n}{(C+1) \ln n \left(\frac{\log n}{\lfloor  \log(\g+1) \rfloor}+2\right)}} \right)~.
\end{eqnarray*}

We note that these bounds are of order $(C(T)+1)\sqrt{n\ln^2 n}$,
respectively $\sqrt{\smash[b]{(C+1) n \ln^2 n}}$, only a factor of $O(\sqrt{\ln
  n})$ larger than the ones of optimal order resulting from earlier algorithms \cite{HeWa98,Vov99,GyLiLu08} which have complexity $O(n^2)$ (strictly speaking, the complexity of \cite{HeWa98}  is $O(nN)$, but, when combined with efficient algorithms designed for the base-expert class, only $O(n^2)$ complexity versions are known \cite{GyLiLu08}).
In some applications, such as online quantization \cite{GyLiLu08}, the number of base experts $N$ depends on the time horizon $n$ in a polynomial fashion, 
that is, $N \sim n^\beta$ for some
$\beta>0$. In such cases
the order of the upper bound is not changed; it remains still $O((C(T)+1) \sqrt{n \ln^2 n})$ if the number
of switches is unknown, and $O(\sqrt{\smash[b]{(C(T)+1)n\ln^2 n}})$ if
the maximum number of switches $C(T)$ is known in advance.  
This bound is within a factor of $O(\sqrt{\ln n})$ of the best achievable
regret for this case.

Next we observe that at the price of a slight increase of
computational complexity, regret bounds of the optimal
order can be obtained.  Indeed, setting $\g= 2n^\gamma-1$ for some
$\gamma\in (0,1)$ and 
$\eta_t\equiv\phi\sqrt{\frac{(2+1/\gamma)\ln n}{n}}, \phi>0$
independently of the maximum number of switches,
\begin{eqnarray*}
\lefteqn{ \hL_n -  L_n(T,a)} \\*
 & \le & \sqrt{n (C(T)+1) \ln N \left(\frac{1}{\gamma}+2\right)}  \\*
&&\mbox{} +
\left(\frac{\phi}{8} + \frac{C+1}{\phi}\right)\sqrt{\left(\frac{1}{\gamma}+2 \right)n\ln n}
+O\left( \sqrt{\frac{n}{\ln n}}\right).
\end{eqnarray*}
If $\eta_t$ is optimized for an a priori known 
bound $C \ge C(T)$, then we get
\begin{eqnarray*}
\lefteqn{\hL_n -  L_n(T,a)} \\* & \le & \sqrt{n (C(T)+1) \left(\frac{1}{\gamma}+2\right)}\left(\sqrt{\ln N} + \sqrt{\frac{\ln n}{2}}\right) \\
&&\mbox{}+O\left( \sqrt{\frac{n}{(C+1)\ln n}}\right).
\end{eqnarray*}
These bounds are of the same $O((C(T)+1)\sqrt{n\ln n})$ and,
respectively, $O(\sqrt{(C+1)n\ln n})$ order as the ones achievable with  the
quadratic complexity algorithms \cite{GyLiLu08, KoSi08}, but
the complexity of our algorithm is only $O(n^\gamma \ln n)$ times
larger than that of running $\A$ (which is typically linear in $n$).
Thus, in a sense the complexity of our algorithm can get very close to
linear while guaranteeing a regret of optimal order. (Note however, that
a factor $1/\sqrt{\gamma}$ appears in the regret bounds so setting $\gamma$ very
small comes at a price.)

A similar behavior is observed for exp-concave loss functions.
Indeed, if $\ell$ is exp-concave and $\A$ is the exponentially weighted
average forecaster, then by \eqref{expconcave} the regret of $\A$ is
bounded by  $\rho_{\E}(n)=
\frac{\ln N}{\eta}$. In this case, for $\g = O(1)$, the bound
\eqref{eq:thmexpconcave} becomes
\begin{eqnarray*}
\lefteqn{\hL_n -  L_n(T,a)}\\
& \le & \frac{(C(T)+1)\left(\frac{\log\frac{n}{C(T)+1}}{\lfloor \log (\g +1)\rfloor}+2\right)}{\eta}
\left(\ln N + \ln n\right)+ O(1).
\end{eqnarray*}
which is a  factor of $O(\ln n)$ larger than the existing optimal bounds
of order $O((C(T)+1)\ln n)$
(see \cite{Wil96, HeWa98,ShMe99, KoSi08, KoRo08}) valid for algorithms
having complexity $O(n^2)$ (again, concerning \cite{HeWa98}, we mean its combination with some efficient algorithm designed for the base-expert class). Note that in this case the algorithm is  
strongly sequential as its parametrization is independent of the time
horizon $n$.
For $\g= 2n^\gamma-1$, we obtain a  bound of optimal order
$O((C(T)+1)\ln n)$:
\begin{eqnarray*}
\lefteqn{\hL_n -  L_n(T,a)} \\ & \le & \frac{(C(T)+1)\left(\frac{1}{\gamma}+2\right)}{\eta}
\left(\ln N + \ln n\right)+ O(1).
\end{eqnarray*}

Bounds of similar order can be obtained for exp-concave loss functions
in the more general case when $\E$ is not of size $N$, but is a bounded convex subset of an
$N$ dimensional linear space. Then $\rho_\E(n)=O(\ln n)$ for several
algorithms $\A$ under different assumptions. 
This is the case for exp-concave loss
functions when $\A$ performs exponential
weighting over all base experts. 
Using random-walk based sampling from log-concave distributions
(see \cite{LoVe06}),
efficient probabilistic approximations
exist to perform this weighting in many cases.
Exact low complexity implementations, such as the 
Krichevsky-Trofimov estimate for the  logarithmic loss
\cite{KrTr81} (see Example~\ref{ex:KT} below), are however, rare. 
When
additional assumptions are made, e.g., the gradient of the loss
function is bounded, the online Newton step algorithm of
\cite{HaAgKa07} can be applied to achieve logarithmic (standard)
regret against the base-expert class $\E$.
We refer to \cite{Bub11} for a survey.

\subsection{The weight function $w^{KT}$} 
\label{subsection:KT}

In this section we analyze the performance of 
Algorithm~\ref{alg:master} for the case when the ``Krichevsky-Trofimov''
weight function $w^{KT}$ is used. Our analysis is
based on part (ii) of
Lemma~\ref{lem:segmentlength}, following ideas of Willems and Krom
\cite{WiKr97} who only considered the logarithmic loss.  Applying the
weight function $\hw^{KT}$ (derived from $w^{KT}$), this analysis improves the constants
relative to Theorem~\ref{thm:main} for small values of $\g$,
although the resulting bound has a less compact form. Nevertheless, in
some special situations the bounds can be expressed in a simple
form. This is the case for the logarithmic loss, where, for the
special choice $\g= 1$, applying
\eqref{eq:segmentlength-refined1}, the new bound now achieves that of
\cite{WiKr97} proved for the same algorithm. The idea is that in the
proof of Theorem~\ref{thm:main} the concavity of $\rho_\E$ was used to
get simple bounds on sums which are sharp if the segments are of
(approximately) equal length. However,  in our
construction the length of the sub-segments (corresponding
to the same segment of the original transition path), or more
precisely, their lower bounds, grow exponentially according to
\eqref{eq:u-increment}. This makes it possible to improve the upper
bounds in Theorem~\ref{thm:main}. It is interesting to note that the
weight functions $w^{\L_1}$ and $w^{\L_2}$ give better bounds for
$\g= n^\gamma$, where the segment lengths are approximately 
equal, while the large differences in the segment lengths for
$\g= O(1)$ can be exploited by the weight function $w^{KT}$.

To obtain ``almost closed-form'' regret bounds for a general $\rho_\E$, we need the following technical lemma.

\medskip

\begin{lemma}
\label{lem:concave}
Assume $f:[1,\infty)\to(0,\infty)$ is a differentiable function and
  $G\ge 1$. Define $F : [1,\infty) \to [0, \infty)$ by 
\[
F(s)=\int_0^{\frac{\log s}{G}} f\left(\frac{s}{2^{cG}}\right)\ dc
\]
for all $s \ge 1$. Then the second derivative of $F$ is given by
\[
F''(s)=\frac{f'(s)}{s G \ln 2}-\frac{f(s)}{s^2 G \ln 2}.
\]
Therefore, $F$ is concave on $[1,\infty)$ if $sf'(s) \le f(s)$ for all $s \ge 1$.
\end{lemma}

\medskip

\begin{IEEEproof}
First note that, since $2^{c G}=s$ for $c=\frac{\log s}{G}$, Leibniz's
integral rule gives
\begin{eqnarray*}
F'(s)&=&\frac{f(1)}{s G \ln 2} + \int_0^{\frac{\log s}{G}} f'\left(\frac{s}{2^{cG}}\right) 2^{-cG}\ dc \\
&=& \frac{f(1)-f(1)+f(s)}{s G \ln 2}=\frac{f(s)}{s G \ln 2}
\end{eqnarray*}
since
\[
- \frac{\partial}{\partial c} \frac{f\left(s2^{-cG}\right)}{s G \ln 2} = f'\left(s2^{-cG}\right) 2^{-cG}.
\]
Differentiating $F'$ gives the desired result.
\end{IEEEproof}

\medskip

Next we give an improvement of Theorem~\ref{thm:main} for small values
of $\g$. For simplicity, the bounds are only given for the tracking
regret. It is much more complicated to obtain sharp bounds for the adaptive regret, since, similarly to the proof of Theorem~\ref{thm:main}, it would require to lower bound the probability that a new segment is started at some time instant $t$, but here the switch probabilities $p_{KT}(t|t')$, defined in \eqref{eq:pKT}, depend both on $t$ and $t'$, unlike $p_{\L_1}(t|t')$ which only depends on $t$.

\medskip

\begin{theorem}
\label{thm:main-ref}
Assume $\rho_\E(x)$ is differentiable and satisfies $\rho_\E(x) \ge x
\rho'_\E(x)$ for all $x \ge 1$, 
and Algorithm~\ref{alg:master} is run with weight function $\{\hw^{KT}_t\}$.
Let
\begin{eqnarray*}
\lefteqn{S(C,n)} \\ &=&(C+1)\int_0^{\frac{\log\frac{n}{C+1}}{\lfloor \log (\g+1)
    \rfloor}} \rho_\E\left( \frac{n}{C+1} 2^{-c \left\lfloor \log(\g+1) \right\rfloor}\right) \ dc   \\
&& \mbox{}+2 (C+1)\rho_\E\left(\frac{n}{C+1}\right)
\end{eqnarray*}
and
\begin{eqnarray*}
\lefteqn{  \bar{r}_n(C)} \\ 
&\!\!\!= &
\!\!\! \frac{(C+1)\ln 2}{4}\Biggl( \frac{\log^2 \frac{n}{C+1}}{\left\lfloor \log(\g+1) \right\rfloor}  \\*
& & \!\!\!\!\! \mbox{} + \biggl(4+\frac{4}{\left\lfloor \log(\g+1)
  \right\rfloor} \biggl) \log \frac{n}{C+1}
 + \left\lfloor \log(\g+1) \right\rfloor +8 \Biggr).
\end{eqnarray*}
If $\ell$ is convex in its first argument and takes values in the
interval $[0,1]$, and $\eta_{t+1} \le \eta_t$ for $t=1,\ldots,n-1$,
then 
for any $T \in \T_n$ the tracking regret satisfies, for all $n$,
\begin{equation}
\label{eq:thmconvex-ref}
\hL_n -  L_n(T,a) \le
 S(C,n) + \sum_{t=1}^n \frac{\eta_t}{8} + \frac{\bar{r}_n(C)}{\eta_n}.
\end{equation}
On the other hand, if $\ell$ is exp-concave for the value of $\eta$ and
$\eta_t = \eta$ for $t=1,\ldots,n$ in Algorithm~\ref{alg:master}, then
for any $T \in \T_n$ the tracking regret satisfies
\begin{equation}
\label{eq:thmexpconcave-ref}
\hL_n -  L_n(T,a) \le S(C,n) + \frac{\bar{r}_n(C)}{\eta_n}.
\end{equation}
\end{theorem}

\medskip

\begin{IEEEproof}
We proceed similarly to the proof of Theorem~\ref{thm:main} by first 
applying  Lemma~\ref{lem:master}. However, the resulting two terms are now
bounded using Lemma~\ref{lem:segmentlength} (ii) instead of Jensen's
inequality, which allows us to make use of the potentially large
differences in the segment lengths.  

For any transition path $T=(t_1,\ldots,t_C) \in \T_n$ let
$\hT=(\th_1,\ldots,\th_{\hC}) \in \T_n$ denote the transition path
defined by Lemma~\ref{lem:segmentlength} with $\hw_n^{KT}(\hT)>0$.  The
first term of the first upper bound given in Lemma~\ref{lem:master} can be bounded
as follows: for any segment $[t_c,t_{c+1})=[\th_{\hc},\th_{\hc'})$ of
    $T$, Lemma~\ref{lem:segmentlength} (i) and
    \eqref{eq:segmentlength-refined2} yield
\begin{eqnarray*}
\lefteqn{\sum_{i=\hc}^{\hc'-1} \rho_\E(\th_{i+1}-\th_i) } \\*
&\le&
\int_0^{\frac{\log(t_{c+1}-t_c)}{\lfloor \log (\g+1) \rfloor}} \rho_\E\left(\frac{t_{c+1}-t_c}{2^{c \left\lfloor \log(\g+1) \right\rfloor}} \right) \ dc 
+2 \rho_\E(t_{c+1}-t_c).
\end{eqnarray*}
Since the right-hand side of the above inequality is a concave function
of $s=t_{c+1}-t_c$ by Lemma~\ref{lem:concave} and the conditions on
$\rho_{\E}$, Jensen's inequality implies 
\begin{eqnarray}
\lefteqn{\sum_{i=0}^{\hC} \rho_\E(\th_{i+1}\!-\th_i)} \nonumber \\*
&\!\!\!\!\!=& \!\!\!\! \sum_{c=0}^C \sum_{i=\hc}^{\hc'-1} \rho_\E(\th_{i+1}\!-\th_i) \nonumber \\
&\!\!\!\!\! \le  &\!\!\!\!  \sum_{c=0}^C \!\!\left(
\!\int_0^{\frac{\log(t_{c+1}\!-t_c)}{\lfloor \log (\g+1) \rfloor}}
\!\! \rho_\E\!\left(\frac{t_{c+1}\!-t_c}{2^{c \left\lfloor \log(\g+1)
    \right\rfloor}} \right)  dc  
+2 \rho_\E(t_{c+1}\!-t_c)
\right) \nonumber \\
& \!\!\!\!\!\le & \!\!\!\! (C+1)\int_0^{\frac{\log\frac{n}{C+1}}{\lfloor \log (\g+1) \rfloor}} \rho_\E\!\left( \frac{n}{C+1} \cdot 2^{-c \left\lfloor \log(\g+1) \right\rfloor}\right)  dc \nonumber \\
&&\!\!\! \mbox{}+2 (C+1)\rho_\E\!\left(\frac{n}{C+1}\right). \label{eq:rhobound-ref}
\end{eqnarray}
The weight function can be bounded in a similar way. By the standard bound \eqref{KTweight} on the
Krichevsky-Trofimov estimate \cite{Sht87}, we have
\begin{eqnarray}
\lefteqn{\ln \frac{1}{\hw^{KT}_n(\hT)}\le \ln \frac{1}{w^{KT}_n(\hT)}  } \qquad  \nonumber \\
&\le & \sum_{c=0}^{\hC} \left(\frac{1}{2}\ln (\th_{c+1}-\th_{c})+\ln
2\right). \label{eq:KTcum} 
\end{eqnarray}
Applying \eqref{eq:segmentlength-refined} for a segment
$[t_c,t_{c+1})=[\th_{\hc},\th_{\hc'})$ of $T$ yields
\begin{eqnarray*}
\lefteqn{\sum_{i=\hc}^{\hc'-1} \left(\frac{1}{2}\ln (\th_{i+1}-\th_{i})+\ln 2\right)} \nonumber \\*
&\le&
\sum_{i=0}^{\left\lceil \frac{\log(t_{c+1}-t_c)}{\left\lfloor \log(\g+1) \right\rfloor} \right\rceil -1} 
\left(\frac{1}{2} \ln \left( \frac{t_{c+1}-t_c}{2^{i \left\lfloor \log(\g+1) \right\rfloor}} \right) + \ln 2 \right) \nonumber \\*
&&\mbox{}+\frac{1}{2} \ln (t_{c+1}-t_c) + \ln 2  \nonumber \\
&=&
\frac{\ln 2}{2} \left\lceil \frac{\log(t_{c+1}-t_c)}{\left\lfloor \log(\g+1) \right\rfloor} \right\rceil \times \nonumber \\*
&&\mbox{} \times \left(\log(t_{c+1}-t_c)-\frac{\left\lceil\frac{\log(t_{c+1}-t_c)}{\left\lfloor
    \log(\g+1) \right\rfloor} \right\rceil-1}{2} \left\lfloor
\log(\g+1) \right\rfloor +2 \right) \nonumber \\
&&\mbox{}+ \frac{1}{2}\ln(t_{c+1}-t_c) + \ln 2
 \nonumber \\
&\le&
\frac{\ln 2}{4}\Biggl( \frac{\log^2(t_{c+1}-t_c)}{\left\lfloor \log(\g+1) \right\rfloor} + \left\lfloor \log(\g+1) \right\rfloor  + 8\\*
&&\quad \quad\quad\quad\mbox{} + \biggl(4+\frac{4}{\left\lfloor \log(\g+1) \right\rfloor} \biggr) \log(t_{c+1}-t_c) 
 \Biggr) \label{eq:segmentbound}
\end{eqnarray*}
where in the last step we bounded the ceiling function from above and from below, as appropriate.
Furthermore, it is easy to check that the last expression above 
is concave in $s=t_{c+1}-t_c$.  Therefore,
combining it with \eqref{eq:KTcum},  applying Jensen's inequality, we obtain
\[
\ln \frac{1}{\hw^{KT}_n(\hT)}  \le \bar{r}_n(C).
\]
Applying this bound and \eqref{eq:rhobound-ref} in
Lemma~\ref{lem:master} yields the statements of the theorem. 
\end{IEEEproof}

\medskip

We now apply Theorem~\ref{thm:main-ref} to the exponentially weighted
average predictor.  For bounded convex loss functions
we have $\rho_\E(n)=\sqrt{n\ln N}$. Assuming $\g=O(1)$,  
if $\eta_t\equiv\phi \sqrt{\frac{2 \ln 2}{n \left \lfloor\log(\g+1)  \right \rfloor}} \log n, \phi>0$
(i.e., $\eta_t$ is independent of the number of switches $C(T)$), we obtain
\begin{eqnarray*}
\lefteqn{\hL_n -  L_n(T,a)} \\
&\le&  2 \sqrt{(C(T)+1) n \ln N} \left(1 +
\frac{1-\sqrt{\frac{C+1}{n}}}{\left \lfloor\log(\g+1)  \right \rfloor
  \ln 2}\right) \\*
& & \mbox{} +\frac{\phi+\frac{C+1}{\phi}}{4}\log n \sqrt{\frac{n \ln 2}{2 \left
    \lfloor\log(\g+1)  \right \rfloor}} + o\bigl((C+1)\sqrt{n}\,\bigr) .
\end{eqnarray*}
Optimizing $\eta_t$ as a function of an upper bound $C$ on the number of switches yields
\begin{eqnarray*}
\lefteqn{\hL_n -  L_n(T,a) } \\
& \le  &
2 \sqrt{(C(T)+1) n \ln N} \left(1 +
\frac{1-\sqrt{\frac{C+1}{n}}}{\left \lfloor\log(\g+1)  \right \rfloor
  \ln 2}\right) \\*
& & \mbox{} +\sqrt{\frac{(C+1) n \log^2 \frac{n}{C+1} \ln 2}{8 \left
    \lfloor\log(\g+1)  \right \rfloor}} + o\bigl(\sqrt{(C+1)n}\,\bigr) . 
\end{eqnarray*}
Note that if $N=  O(n^\beta)$ for some $\beta>0$, the first term is asymptotically negligible compared to the second in the above bounds. For example, if $\eta$ is set independently of $C$, we obtain
\begin{eqnarray*}
\lefteqn{\hL_n -  L_n(T,a)} \\
&\le& \frac{\phi+\frac{C+1}{\phi}}{4}\log n \sqrt{\frac{n \ln 2}{2 \left \lfloor\log(\g+1)
    \right \rfloor}} + o\bigl((C+1)\sqrt{n}\, \bigr).
\end{eqnarray*}
On the other hand, if $\g=2n^\gamma-1$, the bound becomes
\begin{eqnarray*}
\lefteqn{\hL_n -  L_n(T,a) } \\
&\le & 2 \sqrt{(C(T)+1) n \ln N} \left(1 +
\frac{1-\sqrt{\frac{C+1}{n}}}{\gamma \ln n}\right) \\*
&& \mbox{} 
+ \frac{\phi+\frac{C+1}{\phi}}{8}\sqrt{2n \ln n \left(4+ \gamma+\frac{1}{\gamma}\right)}
+ O\left(\sqrt{\frac{n}{\ln n}}\,\right)
\end{eqnarray*}
when $\eta$ is set independently of $C$.

For exp-concave loss functions we have, for $\g = O(1)$, 
\begin{eqnarray*}
\lefteqn{\hL_n -  L_n(T,a)} \\  
&\le &\frac{C+1}{4\eta } \left(\frac{\log \frac{n}{C+1}}{\left \lfloor\log(\g+1)  \right \rfloor} +2\right)
\left(4 \ln N + \ln \frac{n}{C+1}\right) \\
&& \mbox{} + O(C\ln n)
\end{eqnarray*}
while if $\g=2n^\gamma-1$ we get
\begin{eqnarray*}
\lefteqn{\hL_n -  L_n(T,a)} \\  
&\le& \frac{C+1}{4\eta} \left(4\left(\frac{1}{\gamma}+2\right) \ln N + \left(4+\gamma+\frac{1}{\gamma}\right)\ln n\right) \\
&& \mbox{} + O(C).
\end{eqnarray*}

Note that for both types of loss functions we have a clear improvement
relative to Theorem~\ref{thm:main}, where we used the weight function
$w^{\L_1}$, for the case when $\g=O(1)$. However, no such distinction
can be made for $\g=2n^\gamma-1$. Indeed,  for convex loss functions constant
multiplicative changes in $\eta$ vary the exact form of the factor
$(C+a)/b$, with constants $a,b>0$ in the second term, and,
consequently, the order of the bounds depends on the relative size of
$C$, while, for example, the value of $\eta$ determines the order of
the bounds for exp-concave losses, e.g., constructing the weigh function
$\hw$ from $w^{\L_1}$ is better for $\gamma \ge 1/3$.  Also note that the
above bounds for $\g=3$ and $\g=4$ have improved leading constant 
compared to \cite{GyLiLu08b} and \cite{HaSe07}, respectively.

\section{Randomized prediction}
\label{sec:random}

The results of the previous section may be adapted to the closely 
related model of randomized prediction. In this framework, 
the decision maker plays a repeated game against an adversary as follows:
at each
time instant $t=1,\ldots,n$, the decision maker chooses an action
$I_t$ from 
a finite set, say $\{1,\ldots,N\}$ and, independently, the 
adversary assigns losses $\ell_{i,t}\in [0,1]$ to each action $i=1,\ldots,n$.
The goal of the decision maker is to minimize the cumulative loss
$\wh{L}_n = \sum_{t=1}^n \ell_{I_t,t}$.

Similarly to the previous section, the decision maker may try to
compete with the best sequence of actions that can change actions
a limited number of times. More precisely, the set of base
experts is $\E=\{1,\ldots,N\}$ and  as before, we may
define a meta expert that changes base experts $C$ times by
a transition path $T=(t_1,\ldots,t_C;n)$ and a 
vector of actions $a=(i_0,\ldots,i_{C})$, where
$t_0:=1<t_1<\ldots<t_C<t_{C+1}:=n+1$ and $i_j \in \{1,\ldots,N\}$.
The total loss of the  meta expert
    indexed by $(T,a)$, accumulated during
$n$ rounds, is 
\[
  L_n(T,a) = \sum_{c=0}^C L_{i_c}(t_c,t_{c+1})
\]
with
\[
L_{i_c}(t_c,t_{c+1}) = \sum_{t=t_c}^{t_{c+1}-1} \ell_{i_c,t}~.
\]

There are two differences relative to the setup considered
earlier. First, we do not assume that the loss function satisfies
special properties such as convexity in the first argument (although
we do require that it be bounded).  Second,  we do not assume in the
current setup that the action space is convex, and so a convex
combination of the experts' advice is not possible. On the other hand,
similar results as before can be achieved if the decision maker may randomize
its decisions, and in this section we deal with this situation.

In randomized prediction, before taking an action, the decision maker
chooses a probability distribution $\bp_t$ over $\{1,\ldots,N\}$ (a
vector in the probability simplex $\Delta_N$ in $\R^N$), and 
chooses an action $I_t$ distributed according to $\bp_t$
(conditionally, given the past actions of the decision maker and the
losses assigned by the adversary).

Note that now both $\wh{L}_n$ and $L_n(T,a)$ are random variables not
only because the decision takes randomized decisions but also because
the losses set by the adversary may depend on past randomized choices
of the decision maker. (This model is known as the ``non-oblivious
adversary''.)  We may define the \emph{expected loss} of the decision
maker by
\[
   \ol\ell_t(\bp_t) = \sum_{i=1}^N p_{i,t} \ell_{i,t}
\]
 where
$p_{i,t}$ denotes the $i$-th component of $\bp_t$. 

For details and discussion of this
standard model we refer to \cite[Section 4.1]{CeLu06}.
In particular, since the results presented in Section~\ref{sec:intro} can be 
extended to time-varying loss functions and since $\ol\ell_t$ is a linear (convex) function,
it can be shown that regret bounds of any forecaster in the model of Section \ref{sec:intro}
can be extended to the sequence of loss functions  $\ol\ell_t$. That is, the bounds
can be converted into bounds for the expected regret of a randomized forecaster.
Furthermore, it is shown in \cite[Lemma 4.1]{CeLu06} how such bounds in expectation
can be converted to bounds that hold with high probability.

For example, a straightforward combination of  \cite[Lemma 4.1]{CeLu06} 
and Theorem \ref{thm:main} implies the following. Consider a
prediction algorithm $\A$ defined in the model of Section~\ref{sec:general},
that chooses an action in the decision space $\D=\Delta_N$
and suppose that it satisfies a regret bound of the form
\eqref{eq:univbound} under the loss function 
$\ol\ell_t(\bp_t)$.
Algorithm~\ref{alg:rand} below, which is a variant of
Algorithm \ref{alg:master}, 
converts $\A$ into a forecaster under the randomized model.
At each time instant $t$, the algorithm chooses, in a randomized way, a transition path 
$T=(t_1,\ldots,t_C;t) \in \T_t$,
and uses the distribution $\bp_{\A,t}(\tau_t(T))$ that $\A$
would choose, had it been started at time $\tau_t(T)$, the
time of the last change in the path $T$ up to
time $t$. In the definition of the algorithm
\[
\ol{L}_t(\A,T)=\sum_{c=0}^C \ol{L}_\A(t_c,t_{c+1})
\]
denotes the cumulative expected loss of algorithm $\A$, where
we define $t_0=1$ and $t_{c+1}=t+1$, and
\[
\ol{L}_\A(t_c,t_{c+1})=\sum_{s=t_c}^{t_{c+1}-1} \ol{\ell}_s(\bp_{\A,s}(t_c))
\] 
is the cumulative expected loss suffered by $\A$ in the time interval $[t_c,t_{c+1})$ 
with respect to $\ol{\ell}_s$ for $s\in[t_c,t_{c+1})$.

\begin{algorithm}[h]
\caption{Randomized tracking algorithm.}
\label{alg:rand}
{\bf Input:} Prediction algorithm $\A$, weight function $\{w_t; t=1,\ldots,n\}$, 
learning parameters $\eta_t>0, t=1,\ldots,n$. \\
For $t=1,\ldots,n$ 
choose $T \in \T_t$ according to the distribution
\[
q_t(T)= \frac{w_t(T) e^{-\eta_t \ol{L}_{t-1}(\A,T_{t-1})}}{\sum_{T' \in \T_t} w_t(T') e^{-\eta_t \ol{L}_{t-1}(\A,T'_{t-1})}}~,
\]
choose $\bp_t=\bp_{\A,t}(\tau_t(T))$, and pick $I_t \sim \bp_t$.
\end{algorithm}

\begin{corollary}
Suppose $\ell_{i,t} \in [0,1]$ for all $i=1,\ldots,N$ and $t=1,\ldots,n$, and
$\A$ satisfies \eqref{eq:univbound} with respect to the loss function $\{\ell_t\}$.
Assume Algorithm~\ref{alg:rand} is run with weight function $\{\hw^{\L_1}\}$
for some $\e>0$. Let $\delta \in (0,1)$.
For any $T \in \T_n$, the regret of the algorithm satisfies,
with probability at least $1-\delta$,
\begin{eqnarray*}
\lefteqn{\hL_n -  L_n(T,a) } \\*
& \le &
L_{C(T),n} (C(T)+1) \rho_\E\left( \frac{n}{L_{C(T),n} (C(T)+1)}\right) +\sum_{t=1}^n \frac{\eta_t}{8} \\*
&& \mbox{}+ \frac{r_n\left( L_{C(T),n} (C(T)+1)-1\right)}{\eta_n} + \sqrt{\frac{n}{2}\ln \frac{1}{\delta}}~.
\end{eqnarray*}
where $r_n(C)$ and $L_{C,n}$ are defined as in Theorem~\ref{thm:main}.
\end{corollary}

\medskip

\begin{IEEEproof}
First note that Theorem~\ref{thm:main} can easily be extended to
time-varying loss functions (in fact, Lemma~\ref{lem:master},  and
consequently Theorem~\ref{thm:main}, uses
the bound \eqref{convex2} which allows time-varying
loss functions).  Combining the obtained bound for the expected loss
with  \cite[Lemma~4.1]{CeLu06} proves the corollary.
\end{IEEEproof}

\section{Examples}
\label{sec:examples}

In this section we apply the results of the paper for a few specific examples.

\medskip

\begin{example}[Krichevsky-Trofimov mixtures]
\label{ex:KT}
Assume $\D=\E=(0,1)$ and $\Y=\{0,1\}$, and consider the logarithmic loss
defined as $\ell(p,y)=-\I_{y=1}\ln p - \I_{y=0} \ln (1-p)$.  As
mentioned before, the logarithmic loss is exp-concave with $\eta \le
1$, and hence we choose $\eta=1$.  This loss plays a central role in data
compression. In particular, if a prediction method achieves, on a
particular binary sequence $y^n=(y_1,\ldots,y_n)$, a loss $\hL_n$,
then using arithmetic coding the sequence can be described with at
most $\hL_n + 2$ bits \cite{CoTh91}. We note that the choice of the expert
class $\E=(0,1)$ corresponds to the situation where the sequence
$y^n$ is encoded using an i.i.d.\ coding distribution. Competing against the
expert class $\E=(0,1)$ also has a probabilistic interpretation: it is
equivalent to minimizing the worst case maximum coding
redundancy relative to the class of i.i.d.\ source distributions on
$\{0,1\}^n$.

Let $n_0(t)=\sum_{s=1}^t \I_{y_s=0}$ and $n_1(t)=\sum_{s=1}^t \I_{y_s=1}$ denote the number of $0$s and $1$s in $y^t$, 
respectively. Then the loss of an expert $\theta \in (0,1)$ at time $t$ is
\begin{eqnarray*}
L_{\theta,t}&=& -\ln \left((1-\theta)^{n_0(t)}
\theta^{n_1(t)}\right)\\
&=& \mbox{} -n_0(t) \ln(1-\theta)-n_1(t) \ln \theta
\end{eqnarray*}
which is the negative log-probability assigned to $y^t$ by a
memoryless binary Bernoulli source generating $1$s with probability
$\theta$. The Krichevsky-Trofimov forecaster is an exponentially
weighted average forecaster over all experts $\theta \in \E$ using
initial weights $1/(\pi\sqrt{\theta(1-\theta)})$ (i.e., the
Beta$(1/2,1/2)$ distribution) defined as
\begin{eqnarray*}
p^{KT}_t(y^{t-1})&=&\int_0^1 \frac{e^{-L_{\theta,t-1}}}{\pi\sqrt{\theta(1-\theta)}}\ d\theta \\
&=&\int_0^1 \frac{(1-\theta)^{n_0(t-1)} \theta^{n_1(t-1)}}{\pi\sqrt{\theta(1-\theta)}}\ d\theta.
\end{eqnarray*}
It is well known that $p^{KT}_t$ can be computed efficiently as
$p^{KT}_t(1|y^{t-1}) =1-p^{KT}_t(0|y^{t-1})=\frac{n_1(t-1)+1/2}{t}$. By \cite{Sht87}, the performance
of the Krichevsky-Trofimov mixture forecaster can be bounded as
\[
R_n \le \frac{1}{2}\ln n + \ln 2.
\]

In this framework, a meta expert based on the base expert class $\E$
is allowed to change $\theta\in \E$ a certain number of
times. In the probabilistic interpretation, this  corresponds to
the problem of coding a piecewise i.i.d.\ source
\cite{Wil96, WiKr97, ShMe99, MoJa03,RoEr09}.  If we apply 
Algorithm~\ref{alg:master} to this problem with $\hw^{KT}$, we can improve upon Theorem~\ref{thm:main-ref}
by using $\bar{r}_n(C)$ instead of $S(C,n)$ in the bound (note that
$\bar{r}_n(C)$ was obtained by calculating the  Krichevsky-Trofimov bound
for the transition probabilities), and obtain, for any transition path $T \in \T_n$ and meta expert $(T,a)$ 
\begin{eqnarray*}
\lefteqn{\hL_n - L_n(T,a)}\\
&\le& 2 \bar{r}_n(C(T)) \\
&=& \frac{(C(T)+1) \ln 2}{2} \frac{\log^2\frac{n}{C(T)+1}}{\left\lfloor \log(\g+1)\right\rfloor}
+ O((C(T)+1)\ln n).
\end{eqnarray*}
For $\g=1$, this bound recovers that of \cite{WiKr97} (at least in the leading term), and improves the leading
constant for $\g=3$ and $\g=4$ when compared to \cite{GyLiLu08b} and \cite{HaSe09}, respectively.

On the other hand, for $\g=2n^\gamma-1$, $\gamma>0$, using 
with $\hw^{\L_1}$ in Algorithm~\ref{alg:master}, Theorem~\ref{thm:main-ref}
implies
\[
\hL_n - L_n(T,a) \le \frac{3(C(T)+1)}{2}\left(\frac{1}{\gamma}+2\right) \ln n + O(1).
\]
This bound achieves the optimal $O(\ln n)$ order for any $\gamma>0$;
however, with increased leading constant.  On the negative side, for specific choices of
$\gamma$ our algorithm does not recover the best leading constants known in the literature (partly due to the
common bounding technique for all $\gamma$):
If $\gamma=1/2$, our bound is a constant factor worse than those of \cite{MoJa03} and \cite{RoEr09} which have the same
$O(n^{3/2})$ complexity (disregarding logarithmic factors); on the other hand,
in case $\gamma=1$ our algorithm is identical to the $O(n^2)$ complexity algorithm of
Shamir and Merhav \cite{ShMe99}, and hence an optimal bound can be proved for $\hw^{\L_1}$ (and
for $\hw^{\L_2}$), as done in \cite{ShMe99} achieving
Merhav's lower bound \cite{Mer93}. 
\end{example}

\medskip

\begin{example}[Tracking structured classes of base experts]  
In recent years a significant body of research has been devoted to 
prediction problems in which the forecaster competes with a large but 
structured class of experts. We refer to \cite{CeLu06,TaWa03,HeWa09,KoWaKi10,KaVe03,GyLiLu08,CeLu11,DaHaKa08} for an incomplete but representative list of papers.
A quite general framework that has been investigated is the following:
a base expert is represented by a $d$-dimensional binary vector $v\in \{0,1\}^d$.
Let $\E \subset \{0,1\}^d$ be the class of experts. The decision space $\D$
is the convex hull of $\E$, so the forecaster chooses, at each time instant
$t=1,\ldots,n$, a convex combination 
$\wh{p}_t = \sum_{v\in \E} \pi_{v,t} v \in \D \subset [0,1]^d$.
The outcome space is $\Y= [0,1]^d$ and if the outcome is $y_t \in \Y$, then
the loss of expert $v$ is $\ell(v,y_t)= v^Ty_t$, the standard inner
product of $v$ and $y_t$. The loss of the forecaster 
equals $\ell(\wh{p}_t,y_t)= \sum_{v\in \E} \pi_{v,t} v^Ty_t$. 
\cite{KoWaKi10} introduces a general
prediction algorithm, called ``Component Hedge,'' that achieves a regret
\begin{eqnarray*}
\lefteqn{\sum_{t=1}^n\ell(\wh{p}_t,y_y) - \min_{v\in \E} \sum_{t=1}^n \ell(v,y_t)} \\
&\le & d\sqrt{2Kn \ln (d/K)} + dK \ln (d/K) 
\end{eqnarray*}
where $K= \max_{v\in \E} \|v\|_1$. What makes Component Hedge interesting, apart
from its good regret guarantee, is that for many interesting classes of base
experts it can be calculated in time that is polynomial in $d$, even when 
$\E$ has exponentially many experts. We refer to \cite{KoWaKi10} for a list
of such examples. 
The results of this paper show that we may obtain efficiently computable
algorithms for tracking such structured classes of base experts.
For example, \eqref{eq:thmconvex} of Theorem \ref{thm:main} applies in this case,
with $\rho_\E(n) = d\sqrt{2Kn \ln (d/K)} + dK \ln (d/K)$.
The calculations of Section \ref{subsection:exp} may be easily modified 
for this case in a straightforward manner. 
\end{example}

\medskip

\begin{example}[Tracking the best quantizers]
The problem of limited-delay adaptive universal lossy source coding of
individual sequences has recently been investigated in detail
\cite{LiLu01, WeMe02, GyLiLu04, GyLiLu04a, MaWe06, GyLiLu08, GyNe11}.  
In the widely used model of fixed-rate lossy source coding at
rate $R$,  an infinite sequence of $[0,1]$-valued source symbols
$x_1,x_2,\ldots$ is transformed into a sequence of channel symbols
$y_1,y_2,\ldots$ which take values from the finite channel alphabet
$\{1,2,\ldots,M\}$, $M=2^R$, and these channel symbols are then used
to produce the ($[0,1]$-valued) reproduction sequence $\hx_1,\hx_2,\ldots$. 
The quality of the reproduction is measured by the average distortion
$\sum_{t=1}^n d(x_t,\hx_t)$, where $d$ is some nonnegative bounded
distortion measure. The squared error $d(x,x')=(x-x')^2$ is perhaps the
most popular example.

The scheme
is said to have overall delay at most $\delta$ if there exist
nonnegative integers $\delta_1$ and $\delta_2$ with $\delta_1+\delta_2\le \delta$ such
that each channel symbol $y_n$ depends only on the source symbols
$x_1,\ldots,x_{n+\delta_1}$ and the reproduction $\hx_n$ for the source
symbol $x_n$ depends only on the channel symbols
$y_1,\ldots,y_{n+\delta_2}$.  When $\delta=0$, the scheme is said to have
zero delay.  In this case, $y_n$ depends only on $x_1,\ldots,x_n$, and
$\hx_n$ on $y_1,\ldots,y_n$, so that the encoder produces $y_n$ as soon
as $x_n$ becomes available, and the decoder can produce $\hx_n$ when $y_n$
is received. The natural reference class of codes (experts) in this case is the
set of $M$-level scalar quantizers 
\[
\Q=\left\{Q:[0,1]\to\{c_1,\ldots,c_M\},
\{c_1,\ldots,c_M\} \subset[0,1]\right\}~.
\]
The relative loss with respect to the reference class $\Q$ is known
in this context as the distortion redundancy. For the squared error
distortion,  the best randomized coding methods \cite{GyLiLu04,
  GyLiLu04a,GyNe11}, with linear computational complexity with respect
to the set $\Q$, yield a distortion redundancy of order 
$O(n^{-1/4}\sqrt{\ln n})$. 

The problem of competing with the best time-variant quantizer that can
change the employed quantizer several times (i.e., tracking the best
quantizer), was analyzed in \cite{GyLiLu08}, based on a combination of
\cite{GyLiLu04} and the tracking algorithm of \cite{HeWa98}. There the
best linear-complexity scheme achieves $O((C+1) \ln n/n^{1/6})$
distortion redundancy when an upper bound $C$ on the number of
switches in the reference class is known in advance. On the other
hand, applying our scheme with $\g=O(1)$ in the method of
\cite{GyLiLu08} and using the bounds in Section~\ref{subsection:exp},
gives a linear-complexity algorithm with distortion redundancy
$O((C+1)^{1/2} \ln^{3/4}(n)/n^{1/4})+O((C+1)/(\ln^{1/2}( n)/n^{1/2}))$
if $C$ is known in advance and only slightly worse $O((C+1)^{1/2}
\ln^{3/4} (n)/n^{1/4})+O((C+1) \ln( n)/n^{1/2})$ distortion
redundancy if $C$ is unknown.
When $\g=2n^\gamma-1$, the distortion redundancy for linear complexity
becomes somewhat worse, proportional to $n^{-\frac{1}{2(2+\gamma)}}$
up to logarithmic factors.
\end{example}

\section{Conclusion}

We examined the problem of efficiently tracking large expert classes
where the goal of the predictor is to perform as well as a given
reference class. We considered prediction strategies that compete with
the class of switching strategies
that can segment a given sequence into several blocks, and follow the advice of a different base expert in each block.
We derived a family of
efficient tracking algorithms that, for any prediction algorithm $\A$
designed for the base class, can be implemented with time and space
complexity $O(n^{\gamma} \ln n)$ times larger than that of $\A$,
where $n$ is the time horizon and $\gamma \ge 0$
is a parameter of the algorithm. With $\A$ properly chosen, our
algorithm achieves a regret bound of optimal order for $\gamma>0$, and
only $O(\ln n)$ times larger than the optimal order for $\gamma=0$
for all typical regret bound types we examined.  For example, for
predicting binary sequences with switching parameters, our method
achieves the optimal $O(\ln n)$ regret rate with time complexity
$O(n^{1+\gamma}\ln n)$ for any $\gamma\in (0,1)$.
Linear complexity algorithms that achieve optimal
regret rate for small base expert classes have been shown to exist in
\cite{HeWa98} and \cite{KoRo08}. Our results show that
the optimal rate is achievable with the slightly larger
$O(n^{1+\gamma}\ln n), \gamma>0$, complexity even if the number of
switches is not known in advance and the base expert class is
large. It remains, however, an open question whether the optimal rate
is achievable with a linear complexity algorithm in this case.

\section*{Acknowledgment}

We would like to thank the anonymous reviewers for their comments that
helped improve the presentation of the paper.



\bibliographystyle{IEEEtran}

\end{document}